\documentclass[acmtog]{acmart}
\AtBeginDocument{%
  }

\usepackage{enumitem}
\usepackage{multirow}

\newcommand{\rf}[1]{\textcolor{black}{#1}}

\copyrightyear{2026}
\acmYear{2026}
\setcopyright{cc}
\setcctype{by}
\acmConference[SIGGRAPH Conference Papers '26]{Special Interest Group on Computer Graphics and Interactive Techniques Conference Conference Papers}{July 19--23, 2026}{Los Angeles, CA, USA}
\acmBooktitle{Special Interest Group on Computer Graphics and Interactive Techniques Conference Conference Papers (SIGGRAPH Conference Papers '26), July 19--23, 2026, Los Angeles, CA, USA}
\acmDOI{10.1145/3799902.3811193}
\acmISBN{979-8-4007-2554-8/2026/07}
\begin{document}

\title{MTPano: Multi-Task Panoramic Scene Understanding via Label-Free Integration of Dense Prediction Priors}

\author{Jingdong Zhang}
\orcid{0009-0008-6668-1140}
\affiliation{%
  \institution{Texas A\&M University}
  \city{College Station}
  \state{Texas}
  \country{USA}
}
\email{jdzhang@tamu.edu}

\author{Xiaohang Zhan}
\orcid{0000-0003-2136-7592}
\affiliation{%
  \institution{Adobe}
  \country{USA}
}
\email{xiaohangzhan@outlook.com}

\author{Lingzhi Zhang}
\affiliation{%
  \institution{Adobe}
  \country{USA}
}
\email{lingzzha@adobe.com}

\author{Yizhou Wang}
\orcid{0000-0003-1601-9649}
\affiliation{%
  \institution{Adobe}
  \country{USA}
}
\email{wyzjack990122@gmail.com}

\author{Zhengming Yu}
\orcid{0009-0003-0553-8125}
\affiliation{%
  \institution{Texas A\&M University}
  \city{College Station}
  \state{Texas}
  \country{USA}
}
\email{yuzhengming@tamu.edu}

\author{Jionghao Wang}
\orcid{0009-0002-9683-8547}
\affiliation{%
  \institution{Texas A\&M University}
  \city{College Station}
  \state{Texas}
  \country{USA}
}
\email{jionghao@tamu.edu}

\author{Wenping Wang}
\orcid{0000-0002-2284-3952}
\affiliation{%
  \institution{Texas A\&M University}
  \city{College Station}
  \state{Texas}
  \country{USA}
}
\email{wenping@tamu.edu}

\author{Xin Li}
\orcid{0000-0002-0144-9489}
\affiliation{%
  \institution{Texas A\&M University}
  \city{College Station}
  \state{Texas}
  \country{USA}
}
\email{xinli@tamu.edu}

\renewcommand{\shortauthors}{Zhang et al.}

\begin{abstract}
Comprehensive panoramic scene understanding is critical for immersive applications, yet it remains challenging due to the scarcity of high-resolution, multi-task annotations. While perspective foundation models have achieved success through data scaling, directly adapting them to the panoramic domain often fails due to severe geometric distortions and coordinate system discrepancies. Furthermore, the underlying relations between diverse dense prediction tasks in spherical spaces are underexplored. To address these challenges, we propose MTPano, a robust multi-task panoramic foundation model established by a label-free training pipeline. First, to circumvent data scarcity, we leverage powerful perspective dense priors. We project panoramic images into perspective patches to generate accurate, domain-gap-free pseudo-labels using off-the-shelf foundation models, which are then re-projected to serve as patch-wise supervision. Second, to tackle the interference between task types, we categorize tasks into rotation-invariant (e.g., depth, segmentation) and rotation-variant (e.g., surface normals) groups. We introduce the Panoramic Dual BridgeNet (PD-BridgeNet), which disentangles these feature streams via geometry-aware modulation layers that inject absolute position and ray direction priors. To handle the distortion from equirectangular projections (ERP), we incorporate ERP token mixers followed by a dual-branch BridgeNet for interactions with gradient truncation, facilitating beneficial cross-task information sharing while blocking conflicting gradients from incompatible task attributes. Additionally, we introduce auxiliary tasks (image gradient, edge distance field, point map estimation) to fertilize the cross-task learning process. Extensive experiments demonstrate that MTPano achieves state-of-the-art performance on multiple benchmarks and delivers competitive results against task-specific panoramic specialist foundation models. Code and model are released at \url{https://github.com/Evergreen0929/MTPano}
\end{abstract}


\begin{teaserfigure}
  \includegraphics[width=\textwidth]{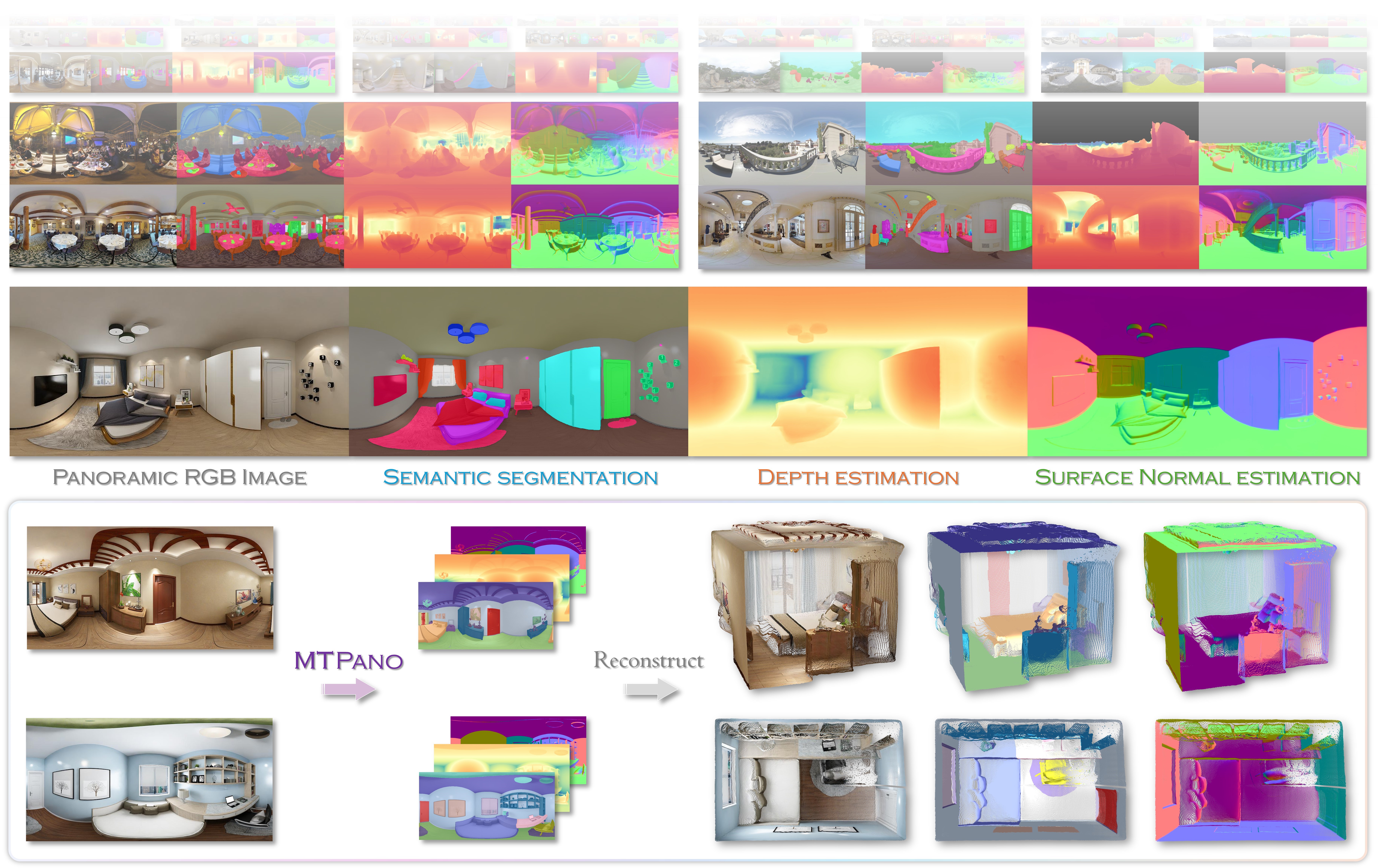}
  \vspace{-6mm}
  \caption{\textbf{Panoramic Scene Understanding via MTPano.} We introduce MTPano, a multi-task foundation model for panoramic dense scene parsing.}
  \label{fig:teaser}
\end{teaserfigure}


\maketitle

\section{Introduction}
\label{sec:intro}
Panoramic scene understanding has become a cornerstone of immersive technologies, empowering applications ranging from Virtual Reality (VR) content creation to ego-centric robot navigation. Unlike narrow field-of-view (FoV) perspective images, $360^{\circ}$ panoramic images offer a holistic observation of the surroundings, necessitating a comprehensive interpretation of geometry (e.g., depth, surface normals) and semantics. Recent years have witnessed remarkable progress in dense prediction tasks~\cite{ye2022inverted,vandenhende2020mti,kirillov2023segment,yang2024depth,yang2024depthv2}, where foundation models have demonstrated exceptional capabilities in parsing complex perspective scenes. This success has naturally catalyzed a surging interest in exploring foundational perception models in the panoramic domain.

Inspired by the progress in the perspective domain, panoramic dense prediction is also witnessing a thriving era, with emerging research striving for high-fidelity spherical scene understanding~\cite{lin2025depth,li20252,huang2024panonormal,zheng2024open,cao2025panda}. However, a major impediment remains: the development of these systems is typically data-driven, yet obtaining high-quality, manually collected pixel-wise annotations for $360^{\circ}$ images is prohibitively expensive and labor-intensive, especially when extended to multiple tasks. To mitigate this data dependency, recent specialists have sought to harness abundant perspective resources. For instance, DA$^2$~\cite{li20252} synthesizes panoramic training data via perspective-to-sphere projection, while PanDA~\cite{cao2025panda} distills priors from perspective foundation models~\cite{yang2024depth} using semi-supervised adaptation. Despite their success, these approaches operate in isolation, treating each modality as a standalone problem. This fragmentation overlooks the critical synergy between tasks (where semantic boundaries can spatially constrain depth discontinuities and vice versa~\cite{vandenhende2020mti}). Furthermore, extending these single-task adaptation strategies to a multi-task setting requires more dedicated designs, as avoiding conflicts and excavating mutual relevances among diverse attributes in a sphere is challenging.

\begin{figure*}[t]
  \centering
  \includegraphics[width=0.99\linewidth]{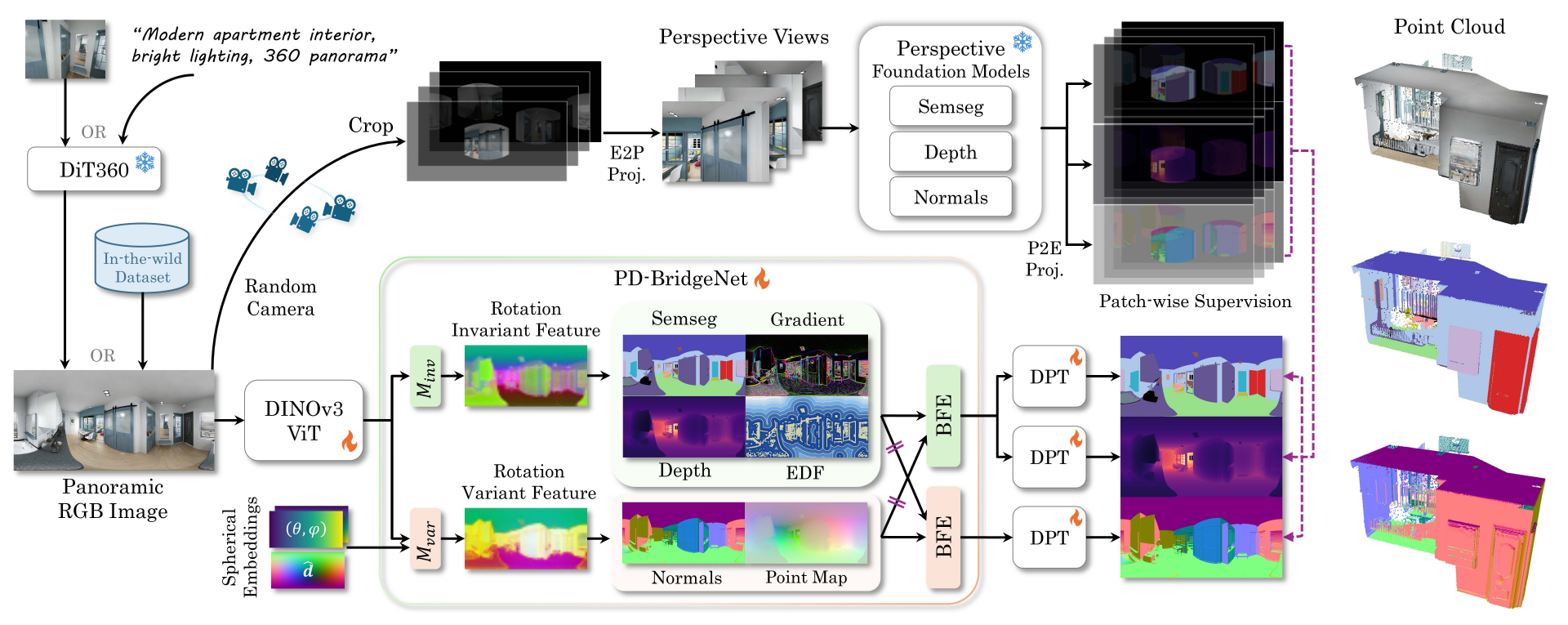}
  \caption{\textbf{Overview of the MTPano framework.} 
    We employ a label-free pipeline (top) that integrates dense priors from perspective foundation models via \textbf{patch-wise supervision}. We propose \textbf{PD-BridgeNet} (bottom), a dual-stream architecture that disentangles rotation-invariant and variant features via geometry-aware modulation ($M_{inv}$ and $M_{var}$). The streams are harmonized by a {Truncated Gradient Flow} mechanism, which facilitates synergistic information exchange while preventing optimization interference across branches. Auxiliary dense task supervisions are involved to aid the task interaction process: Image Gradient, Edge Distance Field (EDF), and Metric Point Map. 
    }
  \label{fig:main}
\end{figure*}

Ideally, Multi-Task Learning (MTL) offers a unified solution to these issues. In the perspective domain, MTL frameworks~\cite{xu2018pad,vandenhende2020mti,ye2022inverted,zhang2025bridgenet} are well-established, primarily focusing on resolving task interference to maximize beneficial feature sharing. However, directly transferring these paradigms to the panoramic domain faces two distinct challenges. First, unlike single-task adaptation, jointly learning diverse dense predictions on a sphere induces severe conflicts. We observe that dense tasks benefit from different priors due to the different responses of coordinate transformations: rotation-invariant tasks (e.g., depth, semantic segmentation) depend solely on relative spatial context and should remain consistent when a transformation is applied, whereas rotation-variant tasks (e.g., surface normals) are strictly tied to the camera coordinates and are orientation-sensitive. Naively sharing features between these conflicting groups leads to \textit{negative transfer}, where cross-task interference leads to mutual degradation. Second, standard perspective MTL architectures are ill-equipped to handle the non-uniform Equirectangular Projection (ERP) distortion. They lack specific mechanisms to model dependencies under ERP distortion while simultaneously decoupling the aforementioned conflicting task attributes. Consequently, simply applying perspective MTL methods to panoramas results in suboptimal performance.

To address these challenges, we introduce MTPano, a label-free framework that establishes a unified multi-task panoramic foundation model by taming dense perspective priors. Our approach tackles the aforementioned hurdles from both data and model perspectives. At the data level, we circumvent the need for manual annotation by integrating knowledge from multiple off-the-shelf perspective foundation models. We project the panoramic image into multiple perspective patches to obtain distortion-free pseudo-labels and then re-project them back as spherical patches for patch-wise supervision. This strategy allows us to leverage the vast knowledge of existing models while preventing overfitting to projection artifacts. At the model level, to resolve the conflict between tasks, we propose the Panorama-Dual-BridgeNet (PD-BridgeNet). Drawing inspiration from BridgeNet~\cite{zhang2025bridgenet}, we design a dual-branch architecture that explicitly disentangles rotation-invariant and rotation-variant feature streams. We employ a distortion-aware {ERP Token Mixer} to handle spherical distortions, and further disentangle the streams by injecting absolute position and ray direction priors into the variant branch via geometry-aware modulation layers. Crucially, we devise an asymmetric bridge mechanism with {Truncated Gradient Flow}: it allows beneficial information to flow between branches while blocking conflicting gradients, ensuring that invariant features are not corrupted by variant supervision and vice versa. Furthermore, we integrate dense auxiliary tasks (including {Image Gradient}, {Edge Distance Field (EDF)}, and {Metric Point Map}) to introduce extra dense priors (e.g. rotation-invariant boundary and texture priors from Image Gradient, coordinate-related geometry priors from metric point maps), facilitating the cross-task learning process.


\par In summary, our contributions are three-fold:
\begin{itemize}
\setlength{\itemsep}{0pt}
\setlength{\parskip}{2pt}
\item We propose MTPano, a multi-task panoramic understanding foundation model training under a label-free pipeline by effectively handling perspective dense priors from strong pretrained perspective specialist models.
\item We identify the critical conflicts between rotation-invariant and -variant features in panoramic MTL and propose PD-BridgeNet, incorporating {geometric feature modulation layers}, distortion-aware token mixing, and a gradient-truncated bridge mechanism to harmonize conflicting features and improve cross-task consistency.
\item Extensive experiments demonstrate that MTPano achieves state-of-the-art performance on multiple benchmarks, and also performs robustly on in-the-wild test cases. 
\end{itemize}

\section{Related Work}

\subsection{Foundational Understanding Models} 
The paradigm of computer vision has significantly shifted from training task-specific models on limited datasets to developing large-scale foundational models driven by massive data and scaling laws. Powered by Vision Transformers~\cite{dosovitskiy2020image} and self-supervised learning~\cite{he2022masked,oquab2023dinov2,simeoni2025dinov3}, these models exhibit exceptional generalization. In segmentation, SAM~\cite{kirillov2023segment} established a new zero-shot standard by extensive pretraining on the SA-1B dataset, while OpenScene~\cite{peng2023openscene} and DINO-X~\cite{ren2024dino} extended open-vocabulary understanding to 3D and generic objects. For geometry estimation, Depth Anything~\cite{yang2024depth,yang2024depthv2} demonstrated the power of data scaling for robust relative depth. To recover accurate metric depth and normals, Metric3D~\cite{yin2023metric3d,hu2024metric3d} and MoGe~\cite{wang2025moge2,wang2025moge} resolved geometric inconsistencies, whereas Marigold~\cite{ke2024repurposing} and Geowizard~\cite{fu2024geowizard} leveraged diffusion priors for state-of-the-art zero-shot inference. These advancements suggest that leveraging such robust representations is a promising pathway to overcome annotation scarcity in specialized domains.

\subsection{Multi-Task Dense Prediction} Multi-Task Learning (MTL) for dense prediction~\cite{gao2019nddr,liu2019end,misra2016cross,vandenhende2020mti,xu2018pad,yang2023contrastive,ye2022inverted,ye2023taskexpert,taskprompter2023,zhang2023rethinking,zhang2018joint,zhang2019pattern,zhang2025bridgenet,tang2025semantic,tian2024unite,lu2024swiss,caomsm,yangmulti,chavhan2025upcycling} aims to learn a unified representation for pixel-wise tasks (e.g., semantic segmentation, depth estimation) to improve efficiency and performance. Early CNN-based approaches focused on decoder-focused feature fusion, such as Cross-Stitch Networks~\cite{misra2016cross} and NDDR-CNN~\cite{gao2019nddr}, which learn linear combinations of task features. To explicitly model task correlations, PAD-Net~\cite{xu2018pad} introduced a multi-modal distillation module to utilize intermediate predictions as priors, while MTI-Net~\cite{vandenhende2020mti} proposed multi-scale task interactions to refine features at different resolutions. With the advent of Transformers, recent works have exploited global context for better task synergy. InvPT~\cite{ye2022inverted} employs an inverted pyramid transformer to model global pixel and task interactions. TaskPrompter~\cite{taskprompter2023} and TaskExpert~\cite{ye2023taskexpert} introduce dynamic prompting and mixture-of-experts mechanisms to disentangle task-specific and task-generic information. More recently, BridgeNet~\cite{zhang2025bridgenet} proposes to leverage bridge features as effective intermediate representations for cross-task interactions, 3D-aware MTL~\cite{li2023multi,wang20253d} extends the multi-task learning to 3D space and multi-view aspect. Beyond deterministic approaches, recent studies have also explored partially supervised MTL settings, such as DiffusionMTL~\cite{ye2024diffusionmtl} and HiTTs~\cite{zhang2025multi}. However, most existing MTL architectures are tailored for perspective images. Directly applying them to panoramic domains fails to address the unique geometric conflicts and distortions inherent in spherical data.

\subsection{Panoramic Understanding Models} Panoramic scene understanding is critical for providing a holistic view of the environment, but suffers from Equirectangular Projection (ERP) distortion and data scarcity. Some works addressed distortion by designing latitude-adaptive window partition~\cite{shen2022panoformer} or utilizing specially designed embeddings like spherical relative positions~\cite{shen2022panoformer} and spherical harmonics~\cite{lee2025hush}. Recently, the focus has shifted towards scaling up data and transferring knowledge from perspective domains. DA$^2$~\cite{li20252} synthesizes panoramic training data via perspective-to-sphere projection and leverages perspective foundation models to generate high-quality pseudo-labels. In parallel, Depth Any Panoramas~\cite{lin2025depth} establishes a metric depth foundation model through a data-in-the-loop paradigm. It constructs a large-scale dataset combining synthetic environments and real-world web data, and employs a DINOv3~\cite{simeoni2025dinov3} backbone with distortion-aware optimization to achieve robust zero-shot generalization without relying on explicit crop-based inference. Other works target specific modalities, such as PanoNormal~\cite{huang2024panonormal} for surface normal estimation and Open Panoramic Segmentation~\cite{zheng2024open} for semantic understanding. Despite these advances, most current methods operate as single-task specialists. The exploration of multi-task panoramic understanding is limited.~\cite{guttikonda2024single} utilizes multi-modal spherical inputs to assist semantic segmentation, yet their primary objective remains restricted to improving a single task rather than achieving holistic scene parsing, while~\cite{shah2024multipanowise,huang2024multi} is one of the few attempts to tackle multiple dense predictions on panoramas, but it relies on standard supervised learning and does not fully leverage the potential of modern foundation models or address the specific invariant \textit{vs.} variant geometric conflicts between tasks.

\section{Method}
\label{sec:method}

In this section, we present \textbf{MTPano}, a unified framework for label-free multi-task panoramic scene understanding. To overcome annotation scarcity and inherent task conflicts, MTPano incorporates two key components: (1) a \textbf{Label-Free Training Pipeline} (Sec.~\ref{sec:pipeline}) that distills dense priors from multiple perspective foundation models via randomized patch-wise supervision; and (2) \textbf{Panorama-Dual-BridgeNet (PD-BridgeNet)} (Sec.~\ref{sec:network}), a dual-stream architecture that disentangles rotation-invariant and variant features via geometric modulation while enabling stable cross-task interaction through a gradient-truncated bridge mechanism.

\begin{figure*}[t]
  \centering
  \includegraphics[width=\linewidth]{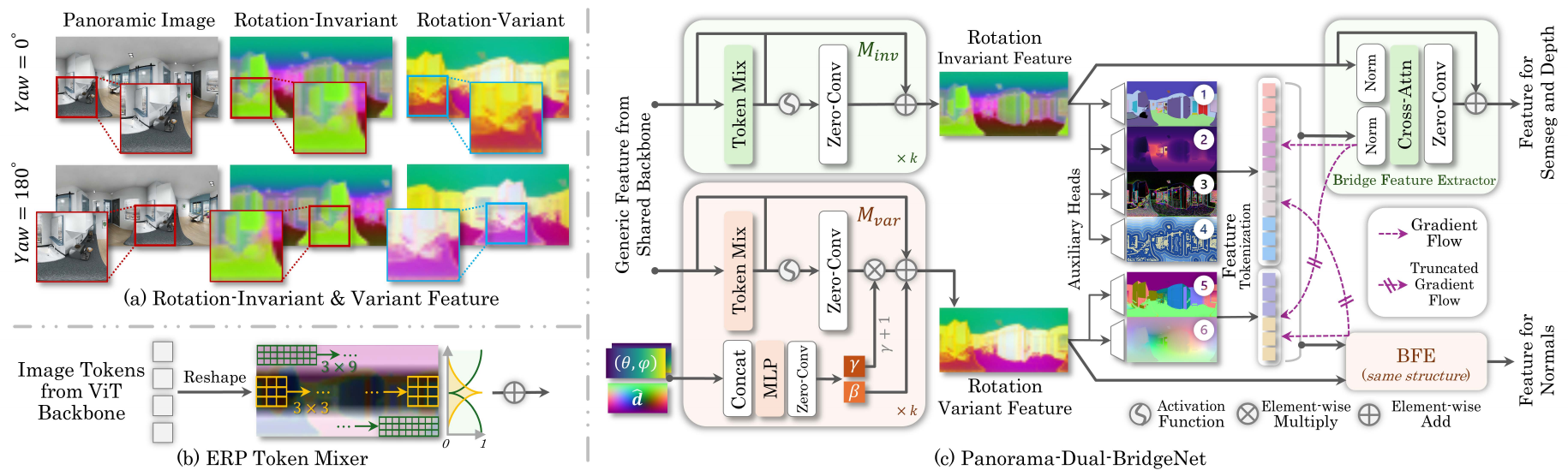}
  \caption{
    \textbf{(a)} We classify dense prediction tasks into rotation-invariant (e.g., Semseg, Depth) and rotation-variant (e.g., Normal) groups based on their dependency on absolute observer orientation. The same region on the rotation-invariant feature remains consistent when rotation on the yaw angle is applied, while the rotation-variant feature doesn't keep this consistency.
    \textbf{(b)} The \textbf{ERP Token Mixer} mitigates spherical distortion by dynamically fusing standard ($3\times3$) and wide ($3\times9$) kernels based on pixel latitude.
    \textbf{(c)} The proposed Panorama-Dual-BridgeNet. We disentangle feature learning into \textbf{Invariant} and \textbf{Variant} Stream via \textbf{Geometry Modulation} layers ($M_{inv}$ and $M_{var}$). The two streams are harmonized by a \textbf{Gradient-Truncated BridgeNet}, which aggregates initial predictions (Semantic Segmentation$^{\textcircled{1}}$, Depth$^{\textcircled{2}}$, Surface Normals$^{\textcircled{5}}$) with dense auxiliary cues (Image Gradient$^{\textcircled{3}}$, Edge Distance Field$^{\textcircled{4}}$, Point Map$^{\textcircled{6}}$) via Cross-Attention to provide thorough interactions while blocking the backward propagation of conflicting gradients.}
  \label{fig:pipeline}
\end{figure*}

\subsection{Label-Free Training Pipeline}
\label{sec:pipeline}



Our pipeline incorporates high-quality data collection/generation and pseudo-annotating from the perspective foundation models. We first collect annotation-free panoramic images from open-source datasets like~\cite{xiao2012recognizing}, in order to provide a higher diversity of data distribution, we also take advantage of current panorama generation methods like~\cite{feng2025dit360,wang2023360} to synthesize a large quantity of indoor/outdoor panoramic scenes.

Despite this abundance of raw data, obtaining high-resolution, pixel-wise multi-task annotations remains prohibitively expensive. As shown in Fig.~\ref{fig:main}, to bypass this bottleneck, we leverage off-the-shelf perspective foundation models by transferring dense priors to the spherical domain via reciprocal projections.
Given an unlabeled panorama $I_{pano}$, we generate $N$ random perspective crops by sampling virtual camera poses with random FoV, yaw $\psi_i$, and pitch $\eta_i$. For each pose, we extract a perspective patch $P_{persp}^i$ using a task-aware P2E projection $\Pi_{P2E}$:
\begin{equation}
    P_{persp}^i = \Pi_{P2E}(I_{pano}, \eta_i, \psi_i) = 
    \begin{cases} 
        I_{pano}(\mathbf{x}_s), & t = {T}_{sem} ,\\
        I_{pano}(\mathbf{x}_s) \cdot (\mathbf{d}_{cam} \cdot \mathbf{k}), & t = {T}_{depth} ,\\
        R(\eta_i, \psi_i)^{-1} \cdot I_{pano}(\mathbf{x}_s), & t = {T}_{norm} ,
    \end{cases}
\label{eq:p2e}
\end{equation}
where $\mathbf{x}_s$ denotes spherical coordinates, $R$ is the rotation matrix, and $\mathbf{d}_{cam} \cdot \mathbf{k}$ accounts for the projection angle relative to the optical axis. Since these patches are distortion-free, we directly apply InternImage-H~\cite{wang2023internimage} and MoGe-2~\cite{wang2025moge2} to obtain high-quality dense predictions $\hat{Y}_{persp}^i$.

Directly stitching these predictions introduces significant artifacts due to scale inconsistencies. Instead, we propose a {Patch-wise Supervision} strategy. We re-project the pseudo-labels back to the spherical coordinate system using the inverse transform $\hat{Y}_{patch}^i = \Pi_{E2P}(\hat{Y}_{persp}^i, \eta_i, \psi_i)$ (i.e., reversing Eq.~\ref{eq:p2e} by dividing depth by $\mathbf{d}_{cam} \cdot \mathbf{k}$ or rotating normals by $R$). During training, we supervise the model using these patches and compute the loss only on valid pixels. This randomized supervision acts as a strong regularization, forcing the network to learn an average distribution consistent across varying views, effectively filtering out projection noise.

\subsection{Panorama-Dual-BridgeNet}
\label{sec:network}
 While the data pipeline provides supervision, standard architectures struggle to handle the inherent ERP distortion and the geometric conflicts between tasks. To address this, we propose the \textbf{Panorama-Dual-BridgeNet (PD-BridgeNet)}. As illustrated in Fig.~\ref{fig:pipeline}(c), our architecture is composed of three key components designed to progressively tame spherical features: (1) a \textbf{Geometric-Aware Disentanglement} module that splits features into rotation-invariant and -variant streams; (2) an \textbf{ERP Token Mixer} that adapts standard ViT features to the non-uniform spherical domain; and (3) a \textbf{Gradient-Truncated Bridge} mechanism that facilitates safe cross-task interaction and avoid negative transfer.

\subsubsection{Geometric-Aware Feature Disentanglement}
\label{sec:disentangle}
\mbox{}\\
\noindent\textbf{Rotation-Invariant vs. Variant Features.}
As illustrated in Fig.~\ref{fig:pipeline} (a), we first categorize dense prediction task features into two groups based on their geometric properties:
\begin{itemize}[leftmargin=*]
\item \textbf{Rotation-Invariant Tasks ($\mathcal{F}_{inv}$):} Tasks like semantic segmentation and depth depend on relative spatial context. Their values describe intrinsic object properties or distances relative to the camera center, which remain consistent regardless of the observer's absolute orientation.
\item \textbf{Rotation-Variant Tasks ($\mathcal{F}_{var}$):} Tasks like surface normal estimation are strictly tied to the absolute coordinate system. Their values change strictly according to the camera's viewing angle.
\end{itemize}
Sharing features naively between these conflicting groups leads to negative transfer. Therefore, we disentangle the feature processing into two parallel streams.

\noindent\textbf{Feature Disentanglement.}
To effectively disentangle the conflicting task features defined above, we construct a dual-stream architecture.
The \textbf{Invariant Stream ($M_{inv}$)} directly employs the ERP Token Mixer (introduced below) to aggregate spatial context. Since tasks in $\mathcal{F}_{inv}$ (e.g., semantic segmentation) rely on intrinsic object properties and relative geometric relations, this stream focuses solely on extracting distortion-free visual representations without introducing directional or absolute positional priors.

In contrast, the \textbf{Variant Stream ($M_{var}$)} depends on absolute spherical coordinates to accurately predict rotation-sensitive attributes. The original features produced by self-supervised pretrained backbones like DINO~\cite{simeoni2025dinov3} are usually rich in consistent semantics but lack perceiving the absolute orientation of the viewer. To bridge this gap, we introduce a set of Spherical Embeddings by explicitly computing the normalized 3D ray direction $\mathbf{d} \in \mathbb{R}^3$ for each pixel. For a spherical coordinate with latitude $\varphi$ and longitude $\theta$, the direction vector components are given by $\mathbf{d} = [\cos\varphi \cos\theta, \sin\varphi, \allowbreak \cos\varphi \sin\theta]^T$.
We concatenate this directional vector $\mathbf{d}$ with the positional embeddings of $(\theta, \varphi)$ to form a condition vector, which is projected via a lightweight MLP to generate pixel-wise affine scale ($\gamma$) and shift ($\beta$) parameters. These parameters dynamically modulate the variant features via the Feature-wise Linear Modulation (FiLM)~\cite{perez2018film} layer: $\mathcal{F}_{var} = (1 + \gamma) \cdot \mathcal{F}_{backbone} + \beta$.
This operation explicitly injects absolute sphere coordinate priors into the feature stream, breaking the spatial invariance and equipping the network with the necessary sense of direction to accurately predict rotation-variant properties like surface normals.

\noindent\textbf{ERP Token Mixer.}
To support dual-stream spherical feature extraction, we must overcome Equirectangular Projection (ERP) distortions, where severe polar stretching degrades standard convolutions. As illustrated in Fig.~\ref{fig:pipeline}(b), we introduce an ERP Token Mixer to extract distortion-aware local features. Specifically, after reshaping the ViT backbone tokens back to the spatial ERP domain, the mixer applies a latitude-adaptive dual-kernel strategy:
\begin{equation}
\text{Mixer}(X) = (1 - w(\varphi)) \cdot (K_{3\times3} * X) + w(\varphi) \cdot (K_{3\times9} * X),
\end{equation}
where $K_{3\times3}$ handles equatorial regions, and the wide $K_{3\times9}$ kernel accommodates polar stretching. The fusion weight $w(\varphi) = |\sin(\varphi)|$ dynamically adjusts the receptive field based on latitude $\varphi$, ensuring uniform spatial aggregation across the sphere.

\begin{table*}[t]
\centering
\caption{Comparisons of panoramic scene understanding on Structured3D, where ${*}$ indicates the method is trained on part of Structured3D training split.}
\label{tab:comparison1}
\setlength{\tabcolsep}{2.3mm}{\scalebox{0.8}{
\begin{tabular}{lccccccccccc}
\toprule
\multirow{2}{*}{Method} & \multicolumn{1}{c}{\textbf{Semseg}} & \multicolumn{5}{c}{\textbf{Depth}} & \multicolumn{5}{c}{\textbf{Normals}} \\
\cmidrule(lr){2-2} \cmidrule(lr){3-7} \cmidrule(lr){8-12}
 & \textit{mIoU} $\uparrow$ & \textit{AbsRel} $\downarrow$ & \textit{RMSE} $\downarrow$ & $\delta_1$ $\uparrow$ & $\delta_2$ $\uparrow$ & $\delta_3$ $\uparrow$ & \textit{Mean} $\downarrow$ & \textit{Median} $\downarrow$ & $<11.5^{\circ}$ $\uparrow$ & $<22.5^{\circ}$ $\uparrow$ & $<30^{\circ}$ $\uparrow$ \\
\midrule
Dinh et al.~\cite{cao2024geometric} & 71.66 & - & - & - & - & - & - & - & - & - & - \\
SFSS-MMSI~\cite{guttikonda2024single} & 71.97 & - & - & - & - & - & - & - & - & - & - \\
$DAP^{*}$~\cite{lin2025depth} & - & 0.0341 & 0.1350 & 98.51 & 99.54 & 99.82 & - & - & - & - & - \\
$DA^{2 *}$~\cite{li20252} & - & 0.0831 & 0.1600 & 95.71 & 98.00 & 98.77 & - & - & - & - & - \\
$PanDA^{*}$~\cite{cao2025panda} & - & 0.0379 & 0.1534 & 98.47 & 99.58 & 99.82 & - & - & - & - & - \\
PanoNormal~\cite{huang2024panonormal} & - & - & - & - & - & - & 5.562 & 0.1048 & 86.68 & 91.01 & 93.08 \\
MultiPanoWise~\cite{shah2024multipanowise} & 69.61 & 0.0557 & 0.1932 & 97.62 & 98.22 & 98.85 & 5.940 & - & - & - & - \\
PanoFormer~\cite{shen2022panoformer} & 64.47 & 0.0861 & 0.3918 & 93.87 & - & - & 7.250 & - & - & - & - \\
HUSH~\cite{lee2025hush} & - & 0.0459 & 0.2577 & 97.76 & 99.22 & 99.61 & 7.014 & 1.6380 & 83.40 & 89.90 & 92.60 \\
InvPT~\cite{ye2022inverted} & 70.02 & 0.0421 & 0.1515 & 97.99 & 98.97 & 99.37 & 5.991 & 0.0822 & 87.81 & 91.18 & 92.71 \\
BridgeNet~\cite{zhang2025bridgenet} & 70.13 & 0.0418 & 0.1492 & 98.17 & 99.01 & 99.42 & 5.988 & 0.0843 & 87.72 & 91.20 & 92.78 \\
TaskPrompter~\cite{taskprompter2023} & 70.95 & 0.0414 & 0.1428 & 98.33 & 99.38 & 99.65 & 5.962 & 0.0727 & 87.88 & 91.44 & 93.02 \\
Ours & \textbf{75.66} & \textbf{0.0248} & \textbf{0.0968} & \textbf{99.27} & \textbf{99.74} & \textbf{99.87} & \textbf{3.850} & \textbf{0.0100} & \textbf{91.66} & \textbf{94.89} & \textbf{96.05} \\
\bottomrule
\end{tabular}
}}
\end{table*}

\begin{table*}[t]
\centering
\caption{Comparisons of panoramic scene understanding on Stanford2D3D.}
\label{tab:comparison2}
\setlength{\tabcolsep}{2.3mm}{\scalebox{0.8}{
\begin{tabular}{lccccccccccc}
\toprule
\multirow{2}{*}{Method} & \multicolumn{1}{c}{\textbf{Semseg}} & \multicolumn{5}{c}{\textbf{Depth}} & \multicolumn{5}{c}{\textbf{Normals}} \\
\cmidrule(lr){2-2} \cmidrule(lr){3-7} \cmidrule(lr){8-12}
 & \textit{mIoU} $\uparrow$ & \textit{AbsRel} $\downarrow$ & \textit{RMSE} $\downarrow$ & $\delta_1$ $\uparrow$ & $\delta_2$ $\uparrow$ & $\delta_3$ $\uparrow$ & \textit{Mean} $\downarrow$ & \textit{Median} $\downarrow$ & $<11.5^{\circ}$ $\uparrow$ & $<22.5^{\circ}$ $\uparrow$ & $<30^{\circ}$ $\uparrow$ \\
\midrule
SFSS-MMSI~\cite{guttikonda2024single}& 60.60 & - & - & - & - & - & - & - & - & - & - \\
Trans4PASS~\cite{zhang2024behind} & 52.10 & - & - & - & - & - & - & - & - & - & - \\
360BEV~\cite{teng2024360bev} & 54.30 & - & - & - & - & - & - & - & - & - & - \\
DAP~\cite{lin2025depth} & - & 0.0686 & 0.4662 & 95.64 & 98.71 & 99.34 & - & - & - & - & - \\
$DA^{2}$~\cite{li20252} & - & \textbf{0.0616} & 0.4261 & \textbf{97.27} & \textbf{98.96} & 99.36 & - & - & - & - & - \\
PanDA~\cite{cao2025panda} & - & 0.0879 & 0.5989 & 93.96 & 98.56 & 99.33 & - & - & - & - & - \\
MultiPanoWise~\cite{shah2024multipanowise} & 54.60 & 0.0649 & 0.3892 & 94.51 & - & - & - & - & - & - & - \\
HUSH~\cite{lee2025hush} & - & 0.0782 & \textbf{0.3332} & 93.84 & 98.49 & 99.29 & 11.191 & 3.8320 & 74.60 & 83.80 & 87.80 \\
Dinh et al.~\cite{cao2024geometric} & 55.50 & 0.1200 & 0.3900 & 86.50 & 98.80 & 99.10 & - & - & - & - & - \\
InvPT~\cite{ye2022inverted} & 56.89 & 0.1207 & 0.5810 & 87.75 & 97.19 & 98.66 & 12.740 & 1.3920 & 76.34 & 82.87 & 85.26 \\
BridgeNet~\cite{zhang2025bridgenet} & 57.12 & 0.1198 & 0.5834 & 87.97 & 97.38 & 98.92 & 12.840 & 1.4650 & 76.29 & 82.67 & 85.15 \\
TaskPrompter~\cite{taskprompter2023} & 57.55 & 0.1171 & 0.5792 & 88.29 & 97.50 & 99.06 & 12.390 & 1.3220 & 76.28 & 82.62 & 85.33 \\
Ours & \textbf{69.47} & 0.0675 & 0.4317 & 96.86 & 98.81 & \textbf{99.37} & \textbf{9.706} & \textbf{0.9314} & \textbf{80.65} & \textbf{86.61} & \textbf{89.11} \\
\bottomrule
\end{tabular}
}}
\end{table*}

\subsubsection{Gradient-Truncated Bridge Mechanism}
\label{sec:bridge}
\mbox{}\\
Independent processing of the two streams guarantees geometric purity but may lose synergistic information. To facilitate safe interaction,  as shown in Fig.~\ref{fig:pipeline} (c), we design a bridge mechanism supported by specific auxiliary supervision. This design serves a dual purpose: auxiliary tasks inject rich geometric and structural cues into the streams, while the truncated bridge ensures these features are shared for interactions without causing optimization conflicts.

\noindent\textbf{Dense Auxiliary Supervision.}
To extract rich task-specific cues for the bridge, we introduce auxiliary tasks for each stream.
For the {Invariant Stream}, we aim to incorporate low-level invariant priors (e.g., textures and boundaries) to aid high-level perception, a strategy validated by~\cite{xu2018pad}. However, direct supervision on sparse edge pixels is inefficient for optimization. Instead, we propose two dense auxiliary tasks: (1) \textbf{Image Gradient Estimation}, which supervises the dense Sobel magnitude and direction; and (2) \textbf{Edge Distance Field (EDF)}, which predicts the unsigned distance field from each pixel to the nearest edge, providing global structural context.
For the {Variant Stream}, to further enforce the understanding of absolute coordinate and global geometry, we introduce estimating \textbf{Metric Point Map}. This task requires the branch to predict the exact dense 3D coordinates $(x, y, z)$ of the scene surface, explicitly aligning the feature space with the spherical spatial distribution.

\noindent\textbf{Bridge Feature Extraction.}
We utilize lightweight auxiliary heads to generate predictions for these tasks. Intermediate features from these heads are extracted and flattened into tokens. To fuse these multi-source cues, we employ {Bridge Feature Extractor (BFE)}~\cite{zhang2025bridgenet} based on Cross-Attention of each stream. The generic backbone features act as \textit{queries}, while the concatenated all task-specific features serve as \textit{keys} and \textit{values}. This aggregates complementary contexts into unified bridge features.

\noindent\textbf{Truncated Gradient Flow.}
A critical challenge in this dual-stream interaction is the conflict between different feature attributes (invariant and variant), which leads to negative transfer during optimization.
To mitigate this, we employ a {Truncated Gradient Flow}. By applying a \texttt{detach} operation to cross-stream features during aggregation, we enable the forward propagation of synergistic context while strictly blocking backward gradients from the opposing branch, effectively isolating optimization interference.





\subsubsection{Optimization Strategy}
\label{sec:optimization}
\mbox{}\\
\noindent\textbf{Zero-Convolution \& Progressive Warmup.}
To effectively leverage the pre-trained DINOv3 backbone, we employ two stabilization strategies. First, we apply \textbf{Zero-Convolution}~\cite{zhang2023adding} to all injection layers. By initializing the injection weights to zero, the network starts as an identity mapping, preserving the foundation model's priors while allowing task-specific refinements to be integrated progressively.
Second, to prevent noise gradients from randomly initialized heads, we adopt a \textbf{Progressive Warmup} strategy. We first freeze the backbone and exclusively optimize the auxiliary heads to align projections with the feature distribution. Once stabilized, we unfreeze the full framework for end-to-end finetuning.

\noindent\textbf{Multi-Task Objective.}
Following standard multi-task optimization protocols~\cite{xu2018pad,ye2022inverted,zhang2025bridgenet}, our total loss $\mathcal{L}_{total}$ is a weighted sum of main task losses and auxiliary supervision losses.
Crucially, to encourage the recovery of fine-grained high-frequency details for geometry tasks (depth and normals), we incorporate the geometry regularization from~\cite{zhang2025spgen} into our objective:
\begin{equation}
    \mathcal{L}_{total} = \sum_{t \in \mathcal{T}_{main}} \lambda_t \mathcal{L}_t + \sum_{t \in \mathcal{T}_{aux}} \lambda_t \mathcal{L}_t +  \sum_{t \in \{{T}_{d}, T_{n}\}} \lambda_{geo}\mathcal{L}_{geo}.
\end{equation}

\section{Experiments}

\subsection{Experimental Settings}

\noindent\textbf{Training Datasets.}
We simulate an unsupervised scenario by compiling a label-free dataset of $\sim$140k panoramas, discarding all original annotations. The collection comprises 20,041 images from Structured3D~\cite{zheng2020structured3d}, 34,260 from Sun360~\cite{xiao2012recognizing}, 10,359 from Matterport3D~\cite{chang2017matterport3d}, and 76,224 synthetic images from DiT360~\cite{feng2025dit360}. For comprehensive spherical coverage, we extract $N=32$ random perspective crops per panorama (FoV $\in [80^{\circ}, 120^{\circ}]$, yaw $\in [0^{\circ}, 360^{\circ}]$, pitch $\in [-90^{\circ}, 90^{\circ}]$). We generate pseudo-labels using perspective foundation models: InternImage-H~\cite{wang2023internimage} for semantic segmentation and MoGe-2~\cite{wang2025moge2} for depth and normals. These predictions are subsequently re-projected onto the sphere as patch-wise supervision for MTPano.

\noindent\textbf{Evaluation Datasets \& Metrics.}
We assess the efficacy of MTPano on two standard indoor panoramic benchmarks: {Structured3D~\cite{zheng2020structured3d}} (synthetic) and {Stanford2D3D~\cite{armeni2017joint}} (real-world). \rf{We further provide more comparisons on Matterport3D~\cite{chang2017matterport3d}, SynPASS~\cite{zhang2024behind}, Deep360~\cite{li2022mode} in our supplementary.} Following the experimental protocols outlined in~\cite{guttikonda2024single,shah2024multipanowise}, we adopt their respective training and testing splits.
Since MTPano inherits the output characteristics of perspective foundation models (e.g., ADE20k semantic categories and MoGe's geometric scale), we finetune MTPano to align the prediction space with the evaluation benchmarks' ground truth for valid comparison.
We report standard dense prediction metrics: mean Intersection over Union (mIoU) for semantic segmentation; Absolute Relative error (AbsRel), RMSE, and $\delta_n$ accuracies for depth; and Mean/Median angular error along with percentage of pixels within angular thresholds ($11.5^{\circ}, 22.5^{\circ}, 30^{\circ}$) for surface normals.

\noindent\textbf{Implementation Details.}
MTPano is implemented in PyTorch using a pre-trained DINOv3-Large~\cite{simeoni2025dinov3} backbone and a DPT~\cite{ranftl2021vision} head (embedding dimension 512) for $512 \times 1024$ inputs. We train the model on 8 NVIDIA A100 GPUs for 100k iterations with a batch size of 4 per GPU. Optimization is performed using Adam (base learning rate $2 \times 10^{-5}$, weight decay $5 \times 10^{-6}$) paired with a polynomial scheduler (power 0.9). Main task losses are weighted equally ($\lambda_{sem}=\lambda_{depth}=\lambda_{norm}=1.0$), with auxiliary losses set to 0.003. To stabilize the dual-stream interaction, we apply our proposed progressive warmup for the first 1,000 steps prior to end-to-end training.


\begin{table}[t]
\centering
\caption{Ablation study on Matterport3D~\cite{chang2017matterport3d}. We use ViT-Small for this experiment. STL for single-task learning and MTL for multi-task learning.}
\label{tab:ablation}
\setlength{\tabcolsep}{3mm}{\scalebox{0.8}{
\begin{tabular}{lcccc}
\toprule
\multirow{2}{*}{Method} & \multicolumn{1}{c}{\textbf{Semseg}} & \multicolumn{1}{c}{\textbf{Depth}} & \multicolumn{1}{c}{\textbf{Normals}} & \multicolumn{1}{c}{\textbf{Delta MTL}} \\
\cmidrule(lr){2-2} \cmidrule(lr){3-3} \cmidrule(lr){4-4} \cmidrule(lr){5-5}
 & \textit{mIoU} $\uparrow$ & \textit{RMSE} $\downarrow$ & \textit{Mean} $\downarrow$ & $\Delta_\mathrm{MTL} (\%)$ \\
\midrule
STL & 24.14 & 0.5523 & 13.3020 & +0.00 \\
MTL baseline & 21.94 & 0.5427 & 13.7740 & -3.64 \\
\textit{w.o.} ERPTokenMixer & 24.08 & 0.5296 & 13.0910 & +1.82 \\
\textit{w.o.} \textit{Trunc} & 24.10 & 0.5300 & 12.9690 & +2.13 \\
\textit{w.o.} $M_{var}$ $M_{inv}$ & 23.09 & 0.5261 & 13.2390 & +0.29 \\
\textit{w.o.} Aux Heads & 24.38 & 0.5278 & 13.1640 & +2.16 \\
PD-BridgeNet & \textbf{26.89} & \textbf{0.5234} & \textbf{12.6370} & \textbf{+7.21} \\
\bottomrule
\end{tabular}
}}
\end{table}

\begin{figure}[t]
  \centering
  \includegraphics[width=\linewidth]{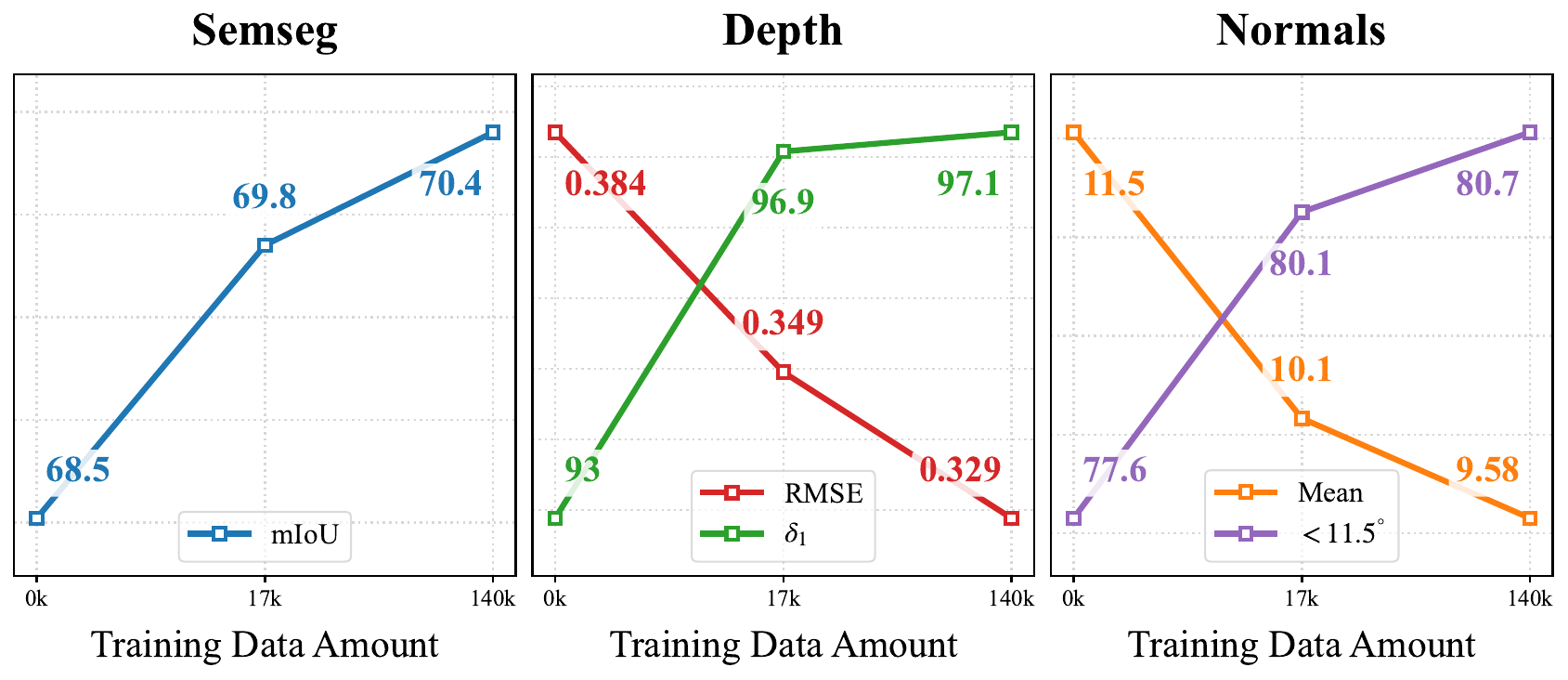}
  \caption{\rf{\textbf{Analysis of data scalability.} We finetune MTPano pretrained with different amounts of our curated data on Stanford2D3D~\cite{armeni2017joint}.}}
  \label{fig:abl_scale}
\end{figure}

\subsection{Qualitative and Quantitative Evaluation}
As shown in Tab.~\ref{tab:comparison1}, on Structured3D, MTPano achieves state-of-the-art performance across all three tasks, consistently outperforming both single-task specialists and multi-task baselines. Specifically, it attains \textbf{75.66\%} mIoU in semantic segmentation, surpassing the previous best SFSS-MMSI~\cite{guttikonda2024single} by \textbf{+3.69\%}. In geometry estimation, our unified model outperforms dedicated specialists, reducing depth AbsRel to \textbf{0.0248} (vs. DAP's 0.0341) and surface normal mean error to \textbf{3.85$^{\circ}$} (vs. PanoNormal's 5.56$^{\circ}$). Notably, compared to standard panoramic MTL baselines like Taskprompter~\cite{taskprompter2023} and BridgeNet~\cite{zhang2025bridgenet}, MTPano reduces depth error by approximately \textbf{40\%}, verifying that our PD-BridgeNet effectively handles task interactions on panoramic images. This performance trend extends to the real-world Stanford2D3D benchmark (Tab.~\ref{tab:comparison2}). MTPano leads semantic segmentation with \textbf{69.47\%} mIoU (\textbf{+8.87\%} gain) and achieves a state-of-the-art normal mean error of \textbf{9.71$^{\circ}$}. In depth estimation, despite a marginal deficit in AbsRel compared to the specialist $DA^{2}$~\cite{li20252}, MTPano still significantly outperforms all multi-task baselines, validating its robust geometric consistency in diverse domains.

\begin{figure*}[t]
  \centering
  \includegraphics[width=\linewidth]{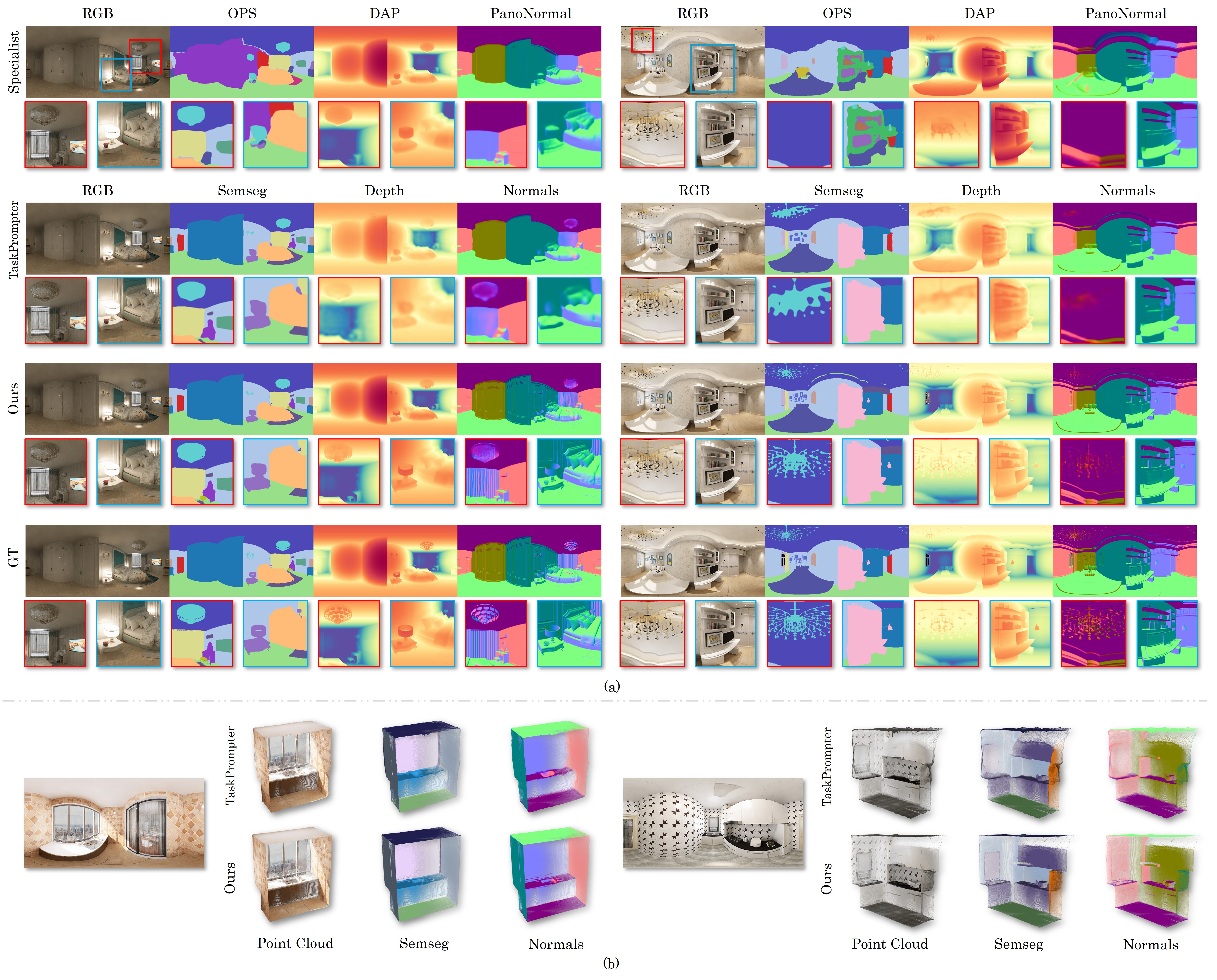}
  \vspace{-4mm}
  \caption{\textbf{Qualitative comparisons on Structured3D.} (a) MTPano outperforms single-task specialists (OPS~\cite{zheng2024open}, DAP~\cite{lin2025depth}, PanoNormal~\cite{huang2024panonormal}) and the multi-task baseline TaskPrompter~\cite{taskprompter2023}, achieving superior segmentation accuracy and geometric detail via PD-BridgeNet's interaction. (b) Point cloud reconstruction comparison demonstrating MTPano's better structural consistency against TaskPrompter.}
  \label{fig:comp_sota}
\end{figure*}

\begin{figure*}[t]
  \centering
  \includegraphics[width=0.99\linewidth]{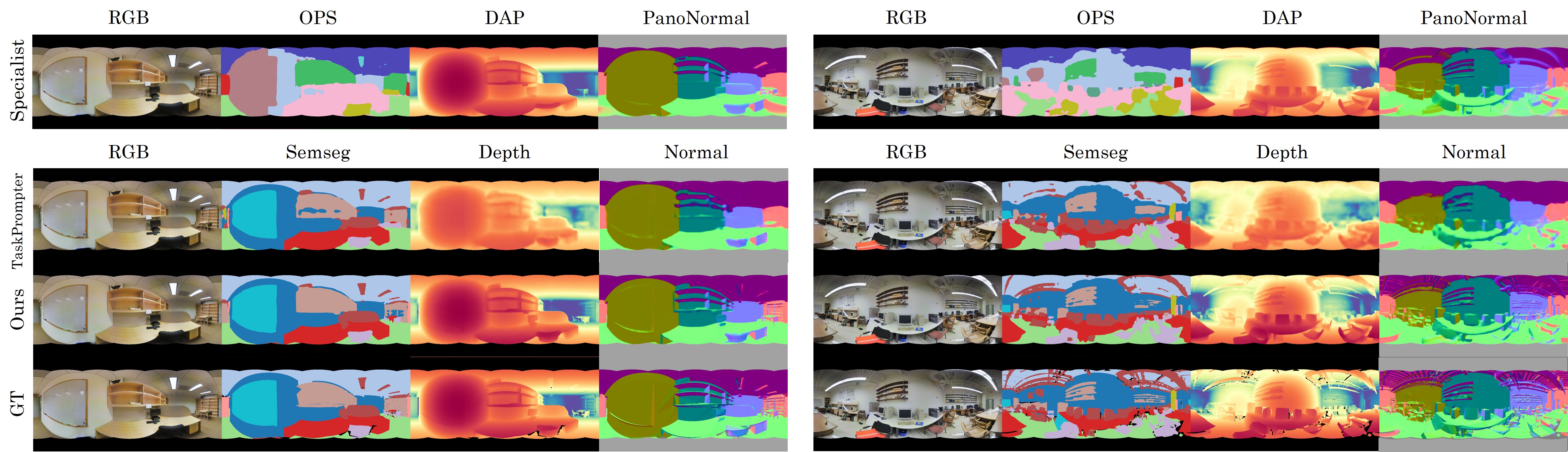}
  \vspace{-1mm}
  \caption{\textbf{Qualitative comparisons with single task specialist models and multi-task models on Stanford2D3D.}}
  \label{fig:comp_sota_2d3d}
\end{figure*}

Qualitative results in Fig.~\ref{fig:comp_sota} demonstrate that MTPano yields significantly sharper predictions than SOTA specialists. Specifically, MTPano captures fine-grained boundaries more accurately compared with DAP~\cite{lin2025depth} and PanoNormal~\cite{huang2024panonormal} (Fig.~\ref{fig:comp_sota}a), and recovers distortion-free 3D point clouds without the geometric noise observed in TaskPrompter~\cite{taskprompter2023} (Fig.~\ref{fig:comp_sota}b). Furthermore, Fig.~\ref{fig:comp_sota_2d3d} validates MTPano's robust generalization on Stanford2D3D dataset, maintaining high semantic and geometric fidelity despite domain gaps.

\subsection{Ablation Study}


\noindent\textbf{Effectiveness of PD-BridgeNet Components.}
\rf{We conduct ablations on Matterport3D~\cite{chang2017matterport3d}. Following~\cite{vandenhende2020mti,zhang2025bridgenet}, we use $\Delta_\mathrm{MTL}$ to measure the average relative improvement over STL baselines. As shown in Tab.~\ref{tab:ablation}, a naive MTL baseline suffers from severe negative transfer ($-3.64\%$). Removing feature disentanglement ($M_{var}$, $M_{inv}$) or the Truncated Gradient degrades performance significantly, confirming that naive feature sharing between conflicting task attributes is suboptimal. Similarly, omitting the ERP Token Mixer or auxiliary heads leads to notable drops. Ultimately, the full PD-BridgeNet effectively resolves these issues, achieving a gain of $\mathbf{+7.21\%}$.}

\noindent\textbf{Data Scalability.} 
\rf{We evaluate MTPano's scalability by finetuning models pretrained on varying volumes of our curated, pseudo-labeled data. As shown in Fig.~\ref{fig:abl_scale}, after finetuning on Stanford2D3D~\cite{armeni2017joint} for just 1k iterations, we observe continuous performance improvements. This demonstrates that purely scaling our hybrid data effectively enhances the model's generalization capabilities.}

\begin{figure*}[t]
  \centering
  \includegraphics[width=\linewidth]{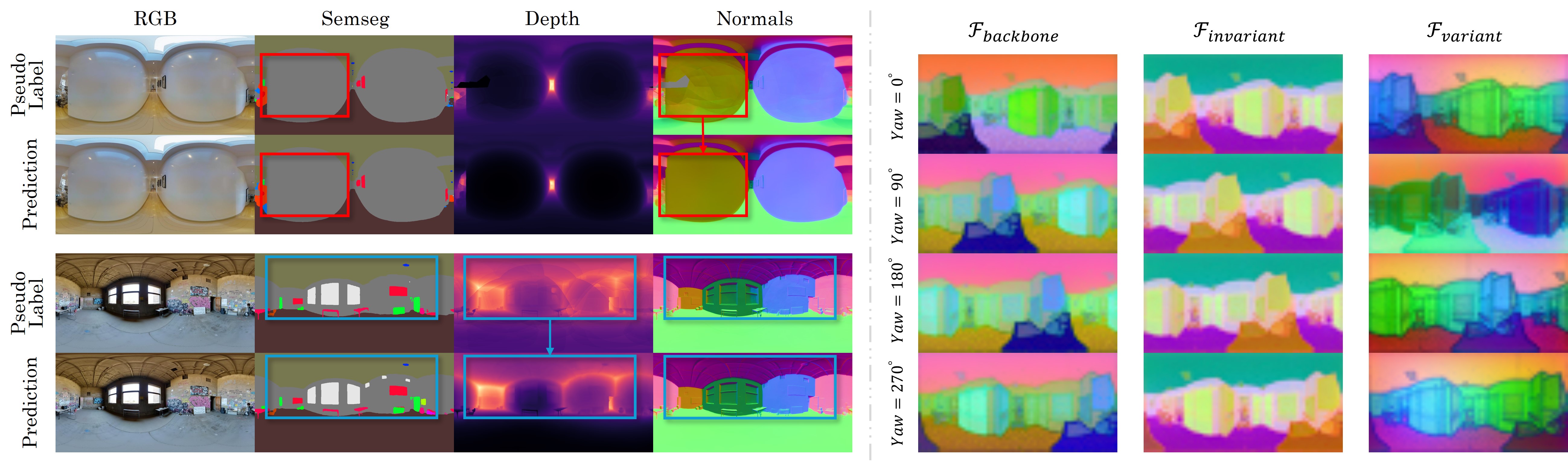}
  \vspace{-5mm}
  \caption{\textbf{Analysis of cross-task learning and feature attributes.} (a) Multi-task interaction effectively eliminates projection artifacts in pseudo-labels. For instance, consistent semantic masks guide surface normal refinement (top), yielding predictions superior to the noisy supervision. (b) Feature visualization under rotation. While backbone features ($\mathcal{F}_{backbone}$) exhibit entangled attributes, our approach successfully disentangles them into rotation-stable invariant features ($\mathcal{F}_{inv}$) and orientation-sensitive variant features ($\mathcal{F}_{var}$).}
  \label{fig:abl}
\end{figure*}

\noindent\textbf{Mutual Correction via Cross-Task Interaction.} 
MTPano effectively mitigates inherent noise and seaming artifacts in pseudo-labels (Fig.~\ref{fig:abl}(a)). By jointly learning complementary tasks, it leverages cross-modal consistency for mutual refinement. For instance, using global semantic context to smooth discontinuous surface normals. Consistent with partially supervised MTL works~\cite{zhang2025multi,li2022learning,ye2024diffusionmtl}, PD-BridgeNet yields predictions visually superior to its noisy supervision, demonstrating effective cross-task fertilization.

\noindent\textbf{Visualizing Feature Disentanglement.} 
Fig.~\ref{fig:abl}(b) illustrates feature behavior under camera rotation. While shared backbone features ($\mathcal{F}_{backbone}$) exhibit entangled attributes, our modulation explicitly decouples them: invariant features ($\mathcal{F}_{inv}$) remain spatially stable to capture high-level semantics, whereas variant features ($\mathcal{F}_{var}$) strictly rotate with the camera to preserve orientation-sensitive cues. This visualization validates the necessity of feature decoupling.

\noindent\rf{\textbf{Model Complexity and Efficiency.}
Compared to STL baselines ($343 \times 3$ MParams, $886 \times 3$ GMACs), MTPano reduces the overall footprint to 590 MParams and 1705 GMACs (PD-BridgeNet accounts for 167 MParams and 340 GMACs). MTPano achieves $161.47 \pm 0.14$ ms inference latency on an A100 GPU.}

\begin{figure*}[t]
  \centering
  \includegraphics[width=\linewidth]{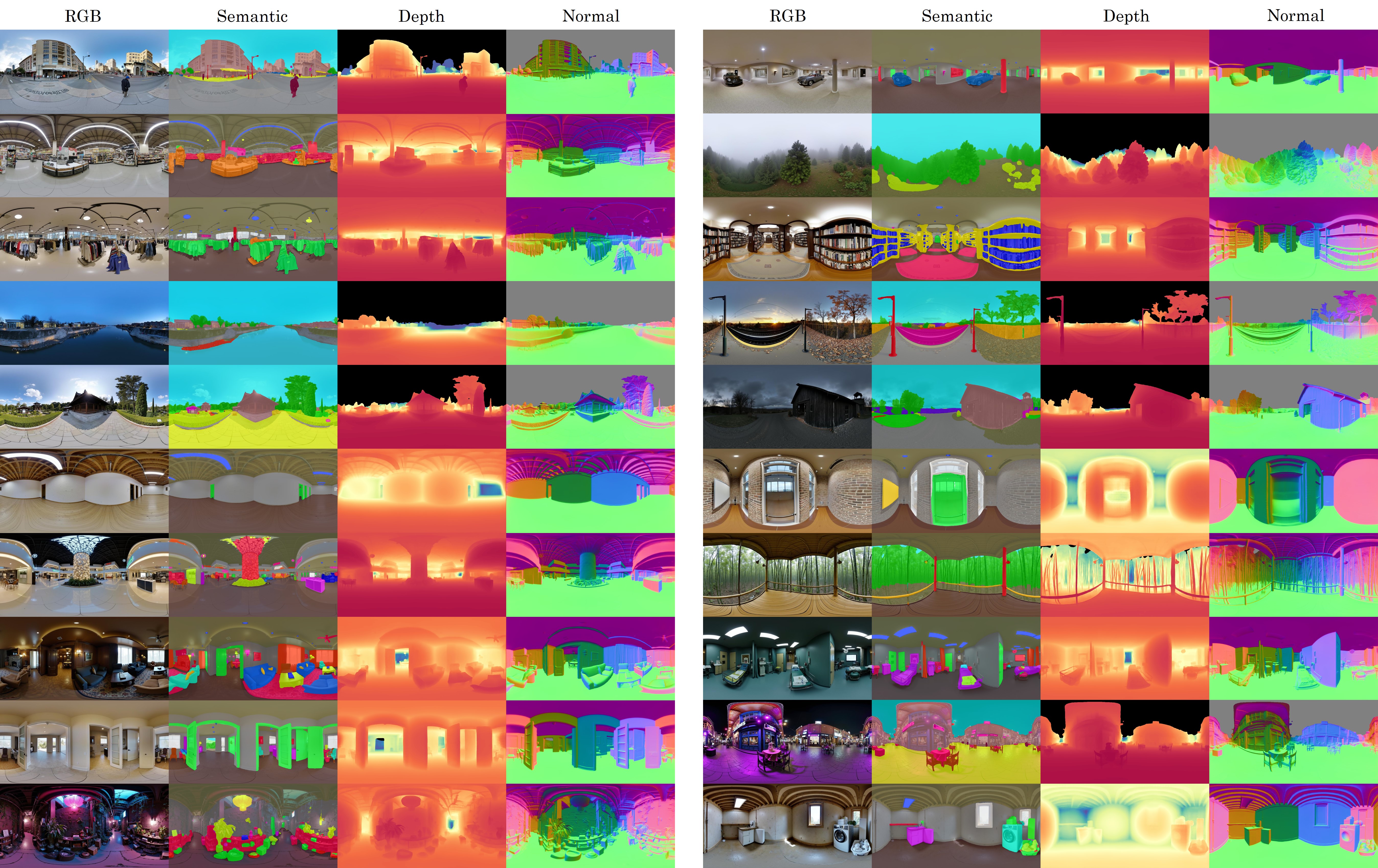}
  \vspace{-3mm}
  \caption{\textbf{Visualization of in-the-wild panoramic scene understanding.}}
  \label{fig:vis_pred}
\end{figure*}

\noindent\textbf{Visualizing In-the-Wild Samples.}
We visualize more samples in Fig.~\ref{fig:vis_pred} generated by DiT360~\cite{feng2025dit360} to test the generalization ability of MTPano. The semantic category is grounded on ADE20k for these samples. \rf{More visualizations, along with additional ablations on task combinations, backbone initializations, and auxiliary task selections, are provided in the supplementary material.}

\section{Conclusion}
We present MTPano, a label-free multi-task foundation model for panoramic scene understanding. By leveraging dense priors from perspective foundation models and introducing the PD-BridgeNet, our framework effectively disentangles rotation-invariant and -variant features while harmonizing their interaction via a truncated gradient mechanism. MTPano achieves state-of-the-art performance on several standard benchmarks, and demonstrates that unified multi-task learning offers a robust and scalable solution for high-fidelity panoramic scene understanding.



\appendix
\vspace{7mm}
\textbf{\Large Supplementary Material}
\vspace{4mm}

In this supplementary material, we provide additional implementation details regarding our data generation pipeline, the specific algorithms used for auxiliary task label generation, \rf{more comparisons on different benchmarks, more ablations on model details and qualitative results demonstrating the generalization capabilities.}

\section{Method Details}

\subsection{Data Generation Pipeline}
\label{sec:data_gen}

To construct a large-scale, diverse, and high-fidelity training dataset without relying on limited real-world captures, we establish a synthetic data generation pipeline utilizing the DiT360~\cite{feng2025dit360} framework. As shown in the provided script, our pipeline consists of two stages: attribute-based prompt engineering and multi-GPU batch generation.

\begin{figure*}[t]
  \centering
  \includegraphics[width=\linewidth]{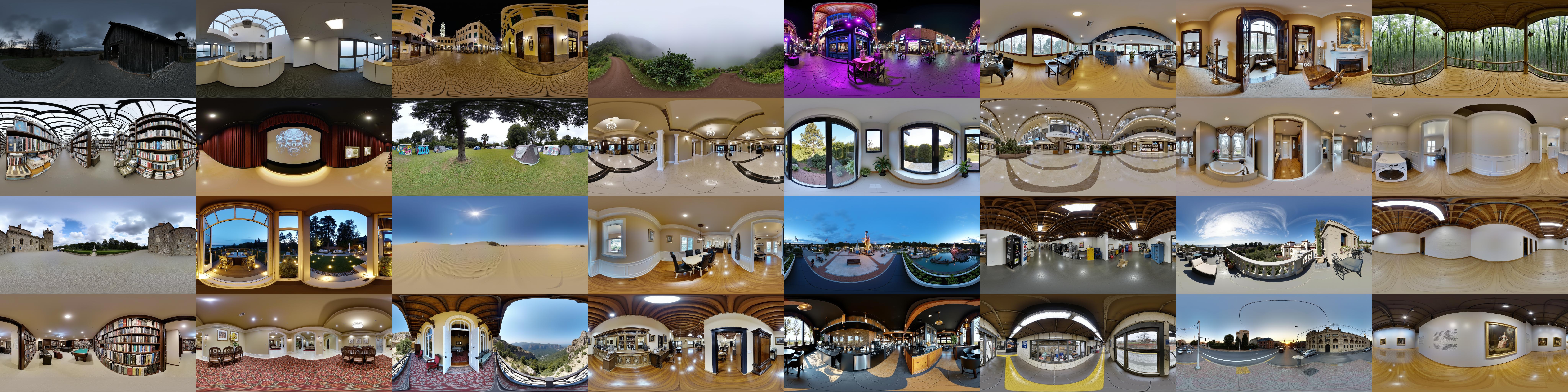}
  \vspace{-5mm}
  \caption{\textbf{Synthetic data generated by DiT360~\cite{feng2025dit360}.}}
  \label{fig:vis_gen}
\end{figure*}

\noindent \paragraph{Attribute-Based Prompt Engineering.} 
To ensure the synthesized dataset covers a wide distribution of scene types and lighting conditions, we construct a comprehensive \textit{Attribute Pool} derived from the distributions of SUN360~\cite{xiao2012recognizing} and Matterport3D~\cite{chang2017matterport3d}. The generation process is governed by a probabilistic template engine:

\begin{itemize}[leftmargin=*]
    \item \textbf{Hierarchical Scene Categorization:} We construct a diverse dictionary of scene attributes categorized into \textit{Indoor} (covering Residential, Commercial, Public Spaces, and Industrial interiors) and \textit{Outdoor} (covering Urban, Nature, Rural, and Historical sites) domains. For each generation task, we sample a scene category (e.g., ``modern kitchen'', ``dense pine forest'') with a weighted probability (60\% Indoor, 40\% Outdoor) to mimic the distribution of real-world applications.
    
    \item \textbf{Probabilistic Modifier Injection:} To prevent visual monotony, we inject three types of modifiers, each with a 50\% independent trigger probability:
    \begin{enumerate}
        \item \textit{Lighting Conditions}: Modifies the global atmosphere (e.g., ``golden hour sunset'', ``cinematic lighting'', ``blue hour twilight'').
        \item \textit{Material \& Texture Details}: Adds fine-grained object descriptions (e.g., ``wooden flooring'', ``exposed brick walls'', ``lush vegetation'').
        \item \textit{Quality Modifiers}: Enforces high-fidelity generation (e.g., ``8k'', ``photorealistic'', ``masterpiece'').
    \end{enumerate}
    
    \item \textbf{Template Construction:} The final prompt $P$ is assembled using a dynamic template structure:
    \begin{equation}
        P = \text{``A } [L] \text{ view of a } [S] \text{ with } [D], [Q], \text{ 360 panorama.''}
    \end{equation}
    where $[S]$ is the mandatory scene description, while $[L]$ (Lighting), $[D]$ (Details), and $[Q]$ (Quality) are optional slots filled based on the Bernoulli trials described above. As shown in Fig.~\ref{fig:vis_gen}, we randomly pick some samples to show the distribution of synthetic data.
\end{itemize}

\noindent \paragraph{Batch Generation.}
We utilize the DiT360 pipeline~\cite{feng2025dit360}, which leverages the FLUX.1-dev model equipped with panorama-specific LoRA adapters (Rank=128) for inference. To maximize throughput for the 140,884 images, we deploy a multi-process distributed generation system on 8$\times$NVIDIA A100 (80GB) GPUs. 
\begin{itemize}[leftmargin=*]
    \item \textbf{Inference Configuration:} We perform generation at a high resolution of $2048 \times 1024$. The denoising process is set to \textbf{35 inference steps} with a \textbf{Guidance Scale (CFG) of 3.0}, which we empirically found to yield the optimal balance between prompt adherence and visual fidelity.
    \item \textbf{Efficiency:} The pipeline operates in \texttt{float16} precision with a batch size of 4 per GPU. We assign a unique random seed to each sample to ensure diversity and employ a distributed task scheduler to handle job allocation and failure recovery robustly.
\end{itemize}

\subsection{Label-Free Training Pipeline}
\label{sec:pipeline}

Our pipeline incorporates high-quality data collection/generation and pseudo-annotating from the perspective foundation models. We first collect annotation-free panoramic images from open-source 

Obtaining high-resolution, pixel-wise multi-task annotations for panoramic images is prohibitively expensive. To bypass this bottleneck, we leverage the rich knowledge encapsulated in off-the-shelf perspective foundation models by transferring dense priors to the spherical domain via reciprocal Perspective-to-Equirectangular (P2E) and Equirectangular-to-Perspective (E2P) projections.

Given an unlabeled panoramic image $I_{pano}$, we generate a set of random perspective crops. Specifically, we sample $N$ virtual camera poses with random Field-of-Views (FoV), yaw $\{\psi_i\}_{1}^{N}$, and pitch $\{\eta_i\}_{1}^{N}$ angles. For each pose, we extract a perspective patch $P_{persp}^i$ using the standard P2E projection $\Pi_{P2E}$:
\begin{equation}
\begin{split}
    P_{persp}^i &= \Pi_{P2E}(I_{pano}, \eta_i, \psi_i) \\
    &= \begin{cases} 
        I_{pano}(\mathbf{x}_s), & \text{if } t = {T}_{sem} ,\\
        I_{pano}(\mathbf{x}_s) \cdot (\mathbf{d}_{cam} \cdot \mathbf{k}), & \text{if } t = {T}_{depth} ,\\
        R(\eta_i, \psi_i)^{-1} \cdot I_{pano}(\mathbf{x}_s), & \text{if } t = {T}_{normal} ,
    \end{cases}
\end{split}
\label{eq:p2e}
\end{equation}
where $\mathbf{x}_s$ denotes the spherical coordinates in the panoramic domain, $R(\eta_i, \psi_i)$ is the rotation matrix determined by the camera yaw $\psi_i$ and pitch $\eta_i$. $\mathbf{d}_{cam} \in \mathbb{R}^3$ indicates the normalized ray direction of $\mathbf{x}_p$ in the local camera coordinate system, and $\mathbf{k} = [0, 0, 1]^T$ represents the principal optical axis unit vector. Since these patches $P_{persp}^i$ are distortion-free, we can directly apply powerful perspective foundation models to generate high-quality pseudo-labels. We utilize InternImage-H~\cite{wang2023internimage} for semantic segmentation and MoGe-2~\cite{wang2025moge2} for geometry estimation (depth and normals), obtaining a set of dense predictions $\hat{Y}_{persp}^i$.

A naive approach would be to stitch these perspective predictions back into a full panoramic pseudo-label map. However, we empirically observe that stitching introduces significant artifacts, since the generated pseudo labels usually contain noise due to scale inconsistencies between overlapping crops, which could cause the model to overfit to the stitching patterns. Instead, we propose a \textbf{Patch-wise Supervision} strategy. We re-project the perspective pseudo-labels back to the spherical coordinate system using the inverse transform $\Pi_{E2P}$:
\begin{equation}
\begin{split}
    \hat{Y}_{patch}^i &= \Pi_{E2P}(\hat{Y}_{persp}^i, \eta_i, \psi_i) \\
    &= \begin{cases} 
        \hat{Y}_{persp}^i(\mathbf{x}_p), & \text{if } t = {T}_{sem} ,\\
        \frac{\hat{Y}_{persp}^i(\mathbf{x}_p)}{\mathbf{d}_{cam} \cdot \mathbf{k}}, & \text{if } t = {T}_{depth} ,\\
        R(\eta_i, \psi_i) \cdot \hat{Y}_{persp}^i(\mathbf{x}_p), & \text{if } t = {T}_{normal} ,
    \end{cases}
\end{split}
\label{eq:e2p}
\end{equation}
where $\mathbf{x}_p$ represents the pixel coordinates in the perspective patch. During training, we supervise the MTPano model using these pseudo label patches directly. We define a valid mask for each patch to compute the loss only on valid pixels. This randomized patch-level supervision acts as a strong regularization: it forces the network to learn an average distribution consistent across varying views, effectively filtering out noise and preventing overfitting to specific projection biases.

\subsection{Dense Auxiliary Supervision}
\label{sec:aux_supervision}

To facilitate cross-task interaction in the PD-BridgeNet, we generate three types of dense auxiliary labels: Image Gradient, Edge Distance Field (EDF), and Metric Point Map. All computations are performed in a fully vectorized manner on the GPU to ensure efficiency.

\paragraph{Image Gradient.}
We compute the image gradient to provide low-level high-frequency cues (e.g., texture and boundaries) for the Rotation-Invariant stream. We first convert the RGB image to grayscale and apply a standard Sobel operator in the tensor space to obtain gradients $G_x$ and $G_y$. The gradient magnitude $M$ and direction $\Phi$ are computed as:
\begin{equation}
    M = \sqrt{G_x^2 + G_y^2}, \quad \Phi = \text{atan2}(G_y, G_x).
\end{equation}
For visualization and supervision, we map the direction $\Phi$ to the Hue channel and the magnitude $M$ to the Value channel in the HSV color space.

\paragraph{Edge Distance Field (EDF).}
The EDF provides global structural context by encoding the distance from each pixel to the nearest edge. We implement this using the \textbf{Jump Flooding Algorithm (JFA)}~\cite{rong2006jump}, which allows for parallel distance transform computation on the GPU. The process is as follows:
\begin{enumerate}
    \item \textbf{Edge Extraction:} We derive a binary edge mask $B$ from the image gradient magnitude using a high threshold ($\tau=0.99$). To prevent the SDF from collapsing at image boundaries, we explicitly clear the border regions of the mask.
    \item \textbf{JFA Distance Transform:} We initialize a seed map where edge pixels store their own coordinates and background pixels store an infinite value. The JFA iteratively propagates the coordinates of the nearest seed with a step size decaying from $N/2$ to $1$. 
    \item \textbf{Distance Calculation:} The final EDF map is obtained by calculating the Euclidean distance between each pixel coordinate and its stored nearest seed coordinate.
\end{enumerate}

\begin{table*}[t]
\centering
\caption{Comparisons of panoramic scene understanding on Matterport3D~\cite{chang2017matterport3d}.}
\vspace{-3.5mm}
\label{tab:comparison3}
\setlength{\tabcolsep}{2.4mm}{\scalebox{0.8}{
\begin{tabular}{lccccccccccc}
\toprule
\multirow{2}{*}{Method} & \multicolumn{1}{c}{\textbf{Semseg}} & \multicolumn{5}{c}{\textbf{Depth}} & \multicolumn{5}{c}{\textbf{Normals}} \\
\cmidrule(lr){2-2} \cmidrule(lr){3-7} \cmidrule(lr){8-12}
 & \textit{mIoU} $\uparrow$ & \textit{AbsRel} $\downarrow$ & \textit{RMSE} $\downarrow$ & $\delta_1$ $\uparrow$ & $\delta_2$ $\uparrow$ & $\delta_3$ $\uparrow$ & \textit{Mean} $\downarrow$ & \textit{Median} $\downarrow$ & $<11.5^{\circ}$ $\uparrow$ & $<22.5^{\circ}$ $\uparrow$ & $<30^{\circ}$ $\uparrow$ \\
\midrule
SFSS-MMSI~\cite{guttikonda2024single} & 35.52 & - & - & - & - & - & - & - & - & - & - \\
Dinh et al.~\cite{cao2024geometric} & 33.06 & - & - & - & - & - & - & - & - & - & - \\
Trans4PASS+~\cite{zhang2024behind} & 29.21 & - & - & - & - & - & - & - & - & - & - \\
DAP~\cite{lin2025depth} & - & 0.1186 & 0.7510 & 85.18 & - & - & - & - & - & - & - \\
$DA^{2}$~\cite{li20252} & - & \textbf{0.0667} & \textbf{0.2882} & \textbf{95.61} & - & - & - & - & - & - & - \\
PanDA~\cite{cao2025panda} & - & 0.0888 & 0.3325 & 92.09 & - & - & - & - & - & - & - \\
360MTL~\cite{huang2024multi} & - & 0.2055 & 0.8446 & 71.29 & 88.63 & 94.77 & 24.658 & 17.6770 & 60.78 & 70.08 & 73.86 \\
MultiPanoWise~\cite{shah2024multipanowise} & 25.52 & 0.1360 & - & 83.30 & 95.29 & - & 20.739 & - & 56.39 & 73.00 & - \\
HUSH~\cite{lee2025hush} & - & - & - & - & - & - & - & - & - & - & - \\
Elite360M~\cite{ai2024elite360m} & 25.38 & 0.1178 & - & 86.72 & 96.27 & - & 17.868 & - & 57.17 & 76.64 & - \\
InvPT~\cite{ye2022inverted} & 27.31 & 0.1145 & - & 87.68 & 96.40 & - & 18.072 & - & 57.00 & 76.19 & - \\
Ours & \textbf{39.11} & 0.1080 & 0.4339 & 93.35 & \textbf{97.49} & \textbf{98.57} & \textbf{9.970} & \textbf{2.7930} & \textbf{80.55} & \textbf{88.68} & \textbf{91.41} \\
\bottomrule
\end{tabular}
}}
\vspace{-2mm}
\end{table*}

\begin{table*}[t]
\centering
\caption{Comparisons of panoramic scene understanding on more benchmarks.}
\vspace{-3.5mm}
\label{tab:comparison4}
\setlength{\tabcolsep}{1.71mm}{\scalebox{0.8}{
\begin{tabular}{lcccccccccccc}
\toprule
\multirow{2}{*}{Method} & \multicolumn{2}{c}{\textbf{Semseg (SynPASS~\cite{zhang2024behind})}} & \multicolumn{5}{c}{\textbf{Depth (Deep360~\cite{li2022mode})}} & \multicolumn{5}{c}{\textbf{Depth (PanoSUNCG~\cite{wang2018self})}} \\
\cmidrule(lr){2-3} \cmidrule(lr){4-8} \cmidrule(lr){9-13}
 & \textit{mIoU} $\uparrow$ (Val.) & \textit{mIoU} $\uparrow$ (Test.) & \textit{AbsRel} $\downarrow$ & \textit{RMSE} $\downarrow$ & $\delta_1$ $\uparrow$ & $\delta_2$ $\uparrow$ & $\delta_3$ $\uparrow$ & \textit{AbsRel} $\downarrow$ & \textit{RMSE} $\downarrow$ & $\delta_1$ $\uparrow$ & $\delta_2$ $\uparrow$ & $\delta_3$ $\uparrow$ \\
\midrule
PVT~\cite{wang2021pyramid} & 37.47 & 32.68 & - & - & - & - & - & - & - & - & - & - \\
SegFormer~\cite{xie2021segformer} & 42.49 & 37.24 & - & - & - & - & - & - & - & - & - & - \\
Trans4PASS+~\cite{zhang2024behind} & 46.61 & 40.72 & - & - & - & - & - & - & - & - & - & - \\
UnmaskFormer~\cite{cao2024occlusion} & 45.34 & - & - & - & - & - & - & - & - & - & - & - \\
UniFuse~\cite{jiang2021unifuse} & - & - & - & - & - & - & - & 0.0528 & 0.2704 & 95.91 & 98.25 & - \\
PanoFormer~\cite{shen2022panoformer} & - & - & - & - & - & - & - & 0.0534 & 0.1890 & 94.87 & 98.83 & - \\
MODE~\cite{li2022mode} & - & - & 0.0365 & - & 97.96 & 99.10 & 99.47 & - & - & - & - & - \\
DAC~\cite{guo2025depth} & - & - & 0.2611 & 8.3710 & 63.11 & - & - & 0.1278 & 0.2788 & 89.67 & 97.85 & - \\
Unik3D~\cite{piccinelli2025unik3d} & - & - & 0.0885 & 6.1480 & 92.93 & - & - & 0.1146 & 0.2538 & 90.18 & 98.02 & - \\
DAP~\cite{lin2025depth} & - & - & 0.0659 & 5.2240 & 95.25 & - & - & - & - & - & - & - \\
PanDA~\cite{cao2025panda} & - & - & - & - & - & - & - & 0.0671 & 0.2185 & 95.42 & 98.25 & - \\
$DA^{2}$~\cite{li20252} & - & - & - & - & - & - & - & 0.0596 & 0.1907 & 96.12 & 98.55 & - \\
Ours & \textbf{49.56} & \textbf{44.71} & \textbf{0.0224} & \textbf{3.1707} & \textbf{98.31} & \textbf{99.36} & \textbf{99.65} & \textbf{0.0312} & \textbf{0.1331} & \textbf{98.53} & \textbf{99.43} & \textbf{99.67} \\
\bottomrule
\end{tabular}
}}
\vspace{-2mm}
\end{table*}

\begin{figure*}[t]
  \centering
  \includegraphics[width=0.99\linewidth]{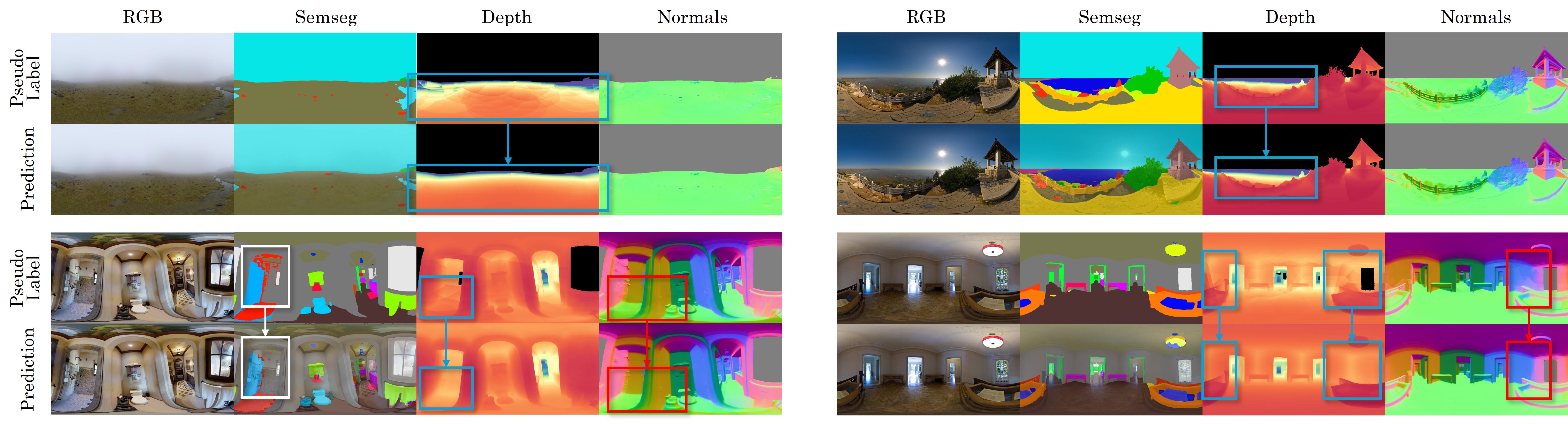}
  \vspace{-3mm}
  \caption{\textbf{More analysis of cross-task learning effects.} Multi-task interaction effectively eliminates projection artifacts in pseudo-labels.}
  \label{fig:abl_gt_more}
\end{figure*}

\begin{figure*}[t]
  \centering
  \includegraphics[width=0.99\linewidth]{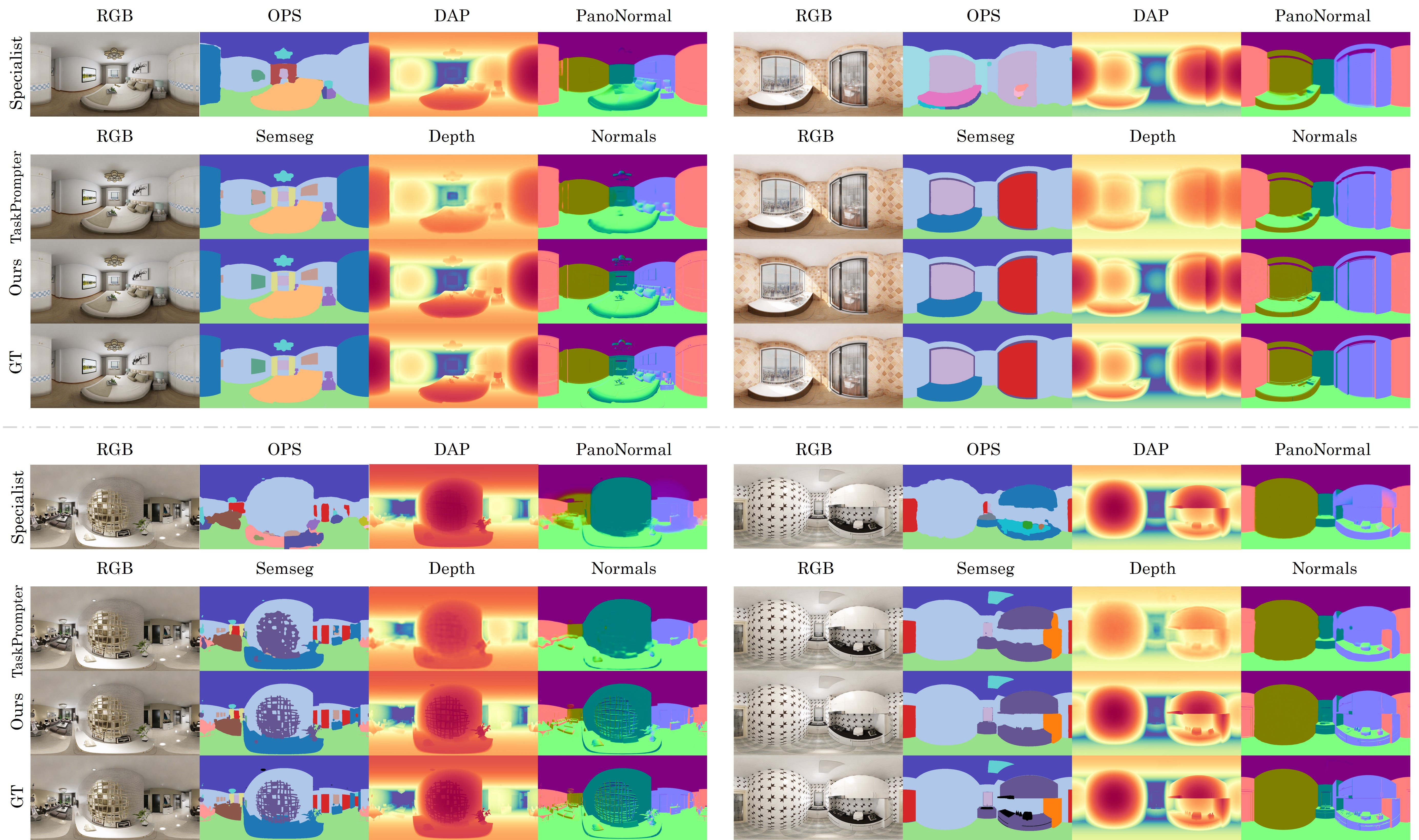}
  \vspace{-3mm}
  \caption{\textbf{More qualitative comparisons with on Structured3D.}}
  \label{fig:vis_comp_more}
\end{figure*}

\paragraph{Metric Point Map.}
To explicitly align the feature space with the spherical spatial distribution for the Rotation-Variant stream, we generate a Metric Point Map. This map represents the absolute 3D coordinate $(x, y, z)$ of the scene surface for each pixel. Given the metric depth map $D \in \mathbb{R}^{H \times W}$ (derived from MoGe-2~\cite{wang2025moge2}), we first generate the spherical unit ray direction map $\mathbf{r} \in \mathbb{R}^{3 \times H \times W}$. For a pixel $(u, v) \in [-1, 1]^2$, the spherical coordinates are $\theta = u\pi$, $\phi = v\frac{\pi}{2}$, and the Cartesian unit vector is:
\begin{equation}
    \mathbf{r}(u, v) = [\cos\phi \sin\theta, \; \sin\phi, \; -\cos\phi \cos\theta]^T.
\end{equation}
The final Metric Point Map is computed as $\mathbf{P} = D \odot \mathbf{r}$. We store these maps in 16-bit integer format (scaled to millimeters) to preserve precision during training.

\begin{table}[t]
\centering
\caption{Ablation study of different task combinations on Matterport3D~\cite{chang2017matterport3d}. We use ViT-Small for this experiment.}
\vspace{-3mm}
\label{tab:ablation_taskcomb}
\setlength{\tabcolsep}{4.6mm}{\scalebox{0.8}{
\begin{tabular}{lccc}
\toprule
\multirow{2}{*}{Method} & \multicolumn{1}{c}{\textbf{Semseg}} & \multicolumn{1}{c}{\textbf{Depth}} & \multicolumn{1}{c}{\textbf{Normals}} \\
\cmidrule(lr){2-2} \cmidrule(lr){3-3} \cmidrule(lr){4-4}
 & \textit{mIoU} $\uparrow$ & \textit{RMSE} $\downarrow$ & \textit{Mean} $\downarrow$ \\
\midrule
STL & 24.14 & 0.5523 & 13.302 \\
Semseg+Depth & 26.41 & 0.5319 & - \\
Semseg+Normals & 26.53 & - & 12.781 \\
Depth+Normals & - & 0.5281 & 12.722 \\
Semseg+Depth+Normals & \textbf{26.89} & \textbf{0.5234} & \textbf{12.637} \\
\bottomrule
\end{tabular}
}}
\vspace{-3mm}
\end{table}

\begin{table}[t]
\centering
\caption{Ablation study of different backbone initilizations on Matterport3D~\cite{chang2017matterport3d}. We use the small-size ViT/DINO for this experiment.}
\vspace{-3mm}
\label{tab:ablation_initlize}
\setlength{\tabcolsep}{1.1mm}{\scalebox{0.8}{
\begin{tabular}{lccc}
\toprule
\multirow{2}{*}{Method} & \multicolumn{1}{c}{\textbf{Semseg}} & \multicolumn{1}{c}{\textbf{Depth}} & \multicolumn{1}{c}{\textbf{Normals}} \\
\cmidrule(lr){2-2} \cmidrule(lr){3-3} \cmidrule(lr){4-4}
 & \textit{mIoU} $\uparrow$ & \textit{RMSE} $\downarrow$ & \textit{Mean} $\downarrow$ \\
\midrule
Random Initialized & 12.76 & 0.6739 & 15.1130 \\
ImageNet~\cite{deng2009imagenet} Pretrained+Warmup & 26.89 & 0.5234 & 12.6370 \\
DINO~\cite{simeoni2025dinov3} Initialized+Warmup & \textbf{28.58} & \textbf{0.4894} & \textbf{12.3870} \\
\bottomrule
\end{tabular}
}}
\vspace{-2mm}
\end{table}

\begin{table}[t]
\centering
\caption{Ablation study on Structured3D~\cite{zheng2020structured3d}. We use DINOv3-Small for this experiment. STL for single-task learning and MTL for multi-task learning.}
\vspace{-3mm}
\label{tab:ablation_s3d}
\label{tab:abl}
\setlength{\tabcolsep}{3mm}{\scalebox{0.8}{
\begin{tabular}{lcccc}
\toprule
\multirow{2}{*}{Method} & \multicolumn{1}{c}{\textbf{Semseg}} & \multicolumn{1}{c}{\textbf{Depth}} & \multicolumn{1}{c}{\textbf{Normals}} & \multicolumn{1}{c}{\textbf{Delta MTL}} \\
\cmidrule(lr){2-2} \cmidrule(lr){3-3} \cmidrule(lr){4-4} \cmidrule(lr){5-5}
 & \textit{mIoU} $\uparrow$ & \textit{RMSE} $\downarrow$ & \textit{Mean} $\downarrow$ & $\Delta_\mathrm{MTL} (\%)$ \\
\midrule
STL & 64.21 & 0.1989 & 5.9120 & +0.00 \\
MTL baseline & 65.17 & 0.1501 & 5.5220 & +10.88 \\
\textit{w.o.} $M_{var}$ $M_{inv}$ \textit{Trunc} & 66.36 & 0.1437 & 5.4850 & +12.77 \\
\textit{w.o.} $M_{var}$ $M_{inv}$ & 66.54 & 0.1434 & 5.4930 & +12.87 \\
\textit{w.o.} Aux Heads & 65.68 & 0.1435 & 5.4850 & +12.46 \\
Full MTPano & \textbf{66.64} & \textbf{0.1432} & \textbf{5.4740} & \textbf{+13.07} \\
\bottomrule
\end{tabular}
}}
\vspace{-2mm}
\end{table}

\begin{table}[t]
\centering
\caption{Ablation study of auxiliary tasks on Matterport3D~\cite{chang2017matterport3d}.}
\vspace{-3mm}
\label{tab:ablation_aux}
\setlength{\tabcolsep}{5.2mm}{\scalebox{0.8}{
\begin{tabular}{lccc}
\toprule
\multirow{2}{*}{Aux. Tasks} & \multicolumn{1}{c}{\textbf{Semseg}} & \multicolumn{1}{c}{\textbf{Depth}} & \multicolumn{1}{c}{\textbf{Normals}} \\
\cmidrule(lr){2-2} \cmidrule(lr){3-3} \cmidrule(lr){4-4}
 & \textit{mIoU} $\uparrow$ & \textit{RMSE} $\downarrow$ & \textit{Mean} $\downarrow$ \\
\midrule
None & 24.38 & 0.5278 & 13.164 \\
+Grad. & 26.28 & 0.5257 & 12.886 \\
+EDF & 25.78 & 0.5266 & 12.892 \\
+Point & 26.01 & \textbf{0.5232} & 12.714 \\
+Point+EDF+Grad. & \textbf{26.89} & {0.5234} & \textbf{12.637} \\
\bottomrule
\end{tabular}
}}
\vspace{-3mm}
\end{table}

\section{More Experimental Results}

\subsection{More Comparisons on Different Benchmarks}

\rf{
\textbf{Comparisons on Matterport3D.} 
In Tab.~\ref{tab:comparison3}, we provide extended comparisons on the challenging real-world Matterport3D~\cite{chang2017matterport3d} dataset. MTPano significantly outperforms previous specialist models (e.g., DAP, 360MTL, Elite360M) across semantic segmentation, depth estimation, and surface normal prediction. Notably, while some baselines (such as DA$^2$ and DAP) involve ground-truth labels and massive training data, our label-free MTPano still achieves superior performance, demonstrating the effectiveness of distilling dense priors from perspective foundation models into a unified multi-task architecture.}

\noindent \rf{
\textbf{Comparisons on Outdoor and Synthetic Benchmarks.} 
To further validate the generalization capability of MTPano, we evaluate it on two additional outdoor benchmarks—SynPASS~\cite{zhang2024behind} for semantic segmentation and Deep360~\cite{li2022mode} for depth estimation—as well as the synthetic PanoSUNCG~\cite{wang2018self} dataset. As shown in Tab.~\ref{tab:comparison4}, MTPano achieves state-of-the-art results across these diverse benchmarks, surpassing specialized models like Trans4PASS+ on SynPASS and DAP on Deep360. This indicates that MTPano generalizes robustly not only to real-world indoor scenes but also to complex outdoor environments.}

\subsection{More Ablations on Model Details}
\rf{
To strictly isolate our architectural contributions and validate our design choices, we conduct further ablation studies on task combinations, backbone initializations, and auxiliary tasks.}

\noindent \rf{
\textbf{Ablation on Task Combinations.} 
As shown in Tab.~\ref{tab:ablation_taskcomb}, we ablate the performance of different task combinations on Matterport3D. Compared to Single-Task Learning (STL) and two-task combinations (e.g., Semseg+Depth or Depth+Normals), the full Semseg+Depth+Normals combination yields the highest performance across all metrics. This proves that MTPano successfully mitigates negative transfer and leverages mutually beneficial cross-task interactions to raise the performance upper bound.}

\noindent \rf{
\textbf{Ablation on Backbone Initialization.} 
In Tab.~\ref{tab:ablation_initlize}, we compare different backbone initialization strategies. Standard backbones typically default to ImageNet~\cite{deng2009imagenet} weights to avoid the slow convergence associated with random initialization. However, initializing with DINOv3~\cite{simeoni2025dinov3} yields the best performance, confirming that inheriting strong visual priors from pre-trained foundation models significantly benefits multi-task panoramic understanding.}

\noindent \rf{
\textbf{Ablation on Architectural Components.} 
In Tab.~\ref{tab:ablation_s3d}, we present the ablation of our proposed PD-BridgeNet components on the Structured3D dataset. Removing the feature disentanglement modules ($M_{var}$, $M_{inv}$), the gradient truncation mechanism (\textit{Trunc}), or the auxiliary heads leads to significant performance drops compared to the full MTL baseline. This validates that our core design, \textit{i.e.} identifying and explicitly resolving rotation-invariant and rotation-variant feature conflicts, is highly effective.}

\noindent \rf{
\textbf{Ablation on Auxiliary Tasks.} 
In the auxiliary tasks ablation (see Tab.~\ref{tab:ablation_taskcomb}), we investigate the impact of individual auxiliary supervision signals. Starting from a baseline with no auxiliary tasks, progressively adding Gradient (Grad.), Edge Distance Field (EDF), and Metric Point Map (Point) steadily improves the performance. The full combination provides the most comprehensive information, yielding better overall performance.
}



\section{More Visualized Results}
We visualize more samples in Fig.~\ref{fig:vis_comp_more} generated by DiT360~\cite{feng2025dit360} to showcasing MTPano's robust performance across semantic segmentation, depth estimation, and surface normal prediction in varied lighting and scene conditions. The semantic category is grounded on ADE20k for these samples.


\begin{figure*}[t]
  \centering
  \includegraphics[width=0.95\linewidth]{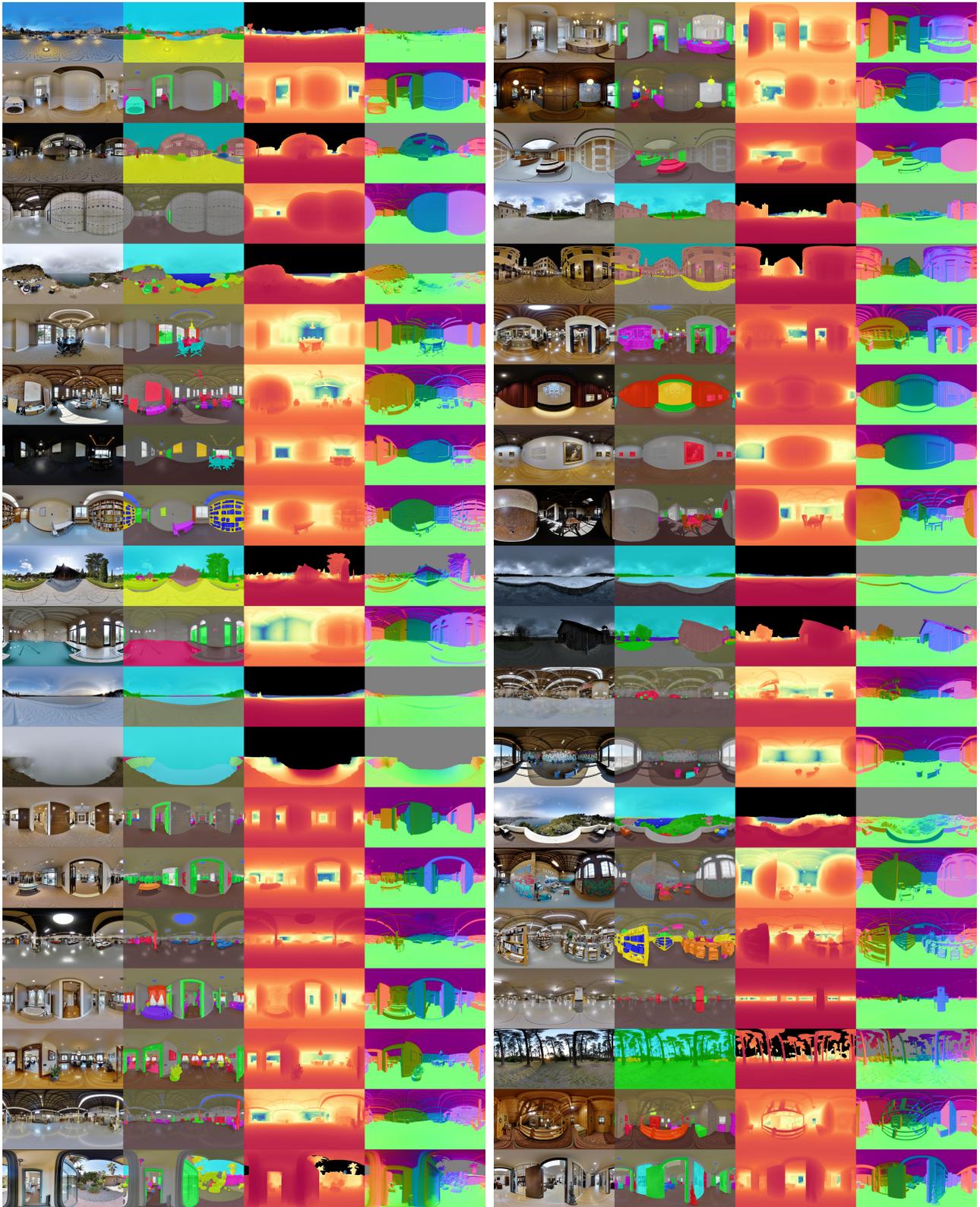}
  \vspace{-3mm}
  \caption{\textbf{More results on in-the-wild data samples.}}
  \label{fig:vis_more}
\end{figure*}

\bibliographystyle{ACM-Reference-Format}
\bibliography{sample-base}

\end{document}